%% file: main.tex
\pdfoutput=1
\documentclass[11pt]{article}

\usepackage[]{acl}

\usepackage{times}
\usepackage{latexsym}
\usepackage{graphicx} % DO NOT CHANGE THIS
\usepackage{subfigure}
\usepackage{amssymb,amsthm,dsfont,mathrsfs}
\usepackage{amsmath}
\usepackage{multirow}
\input{definition}

\usepackage[T1]{fontenc}
\usepackage[utf8]{inputenc}

\usepackage{microtype}
\title{Robust Lottery Tickets for Pre-trained Language Models}
\author{{\normalsize
    Rui Zheng$^{\bigstar*}$, \ \ Rong Bao$^{\bigstar}$\thanks{$^*$  Equal contribution.} , \ \ Yuhao Zhou$^{\bigstar}$, \ \ Di Liang$^{\spadesuit}$, \ \ Sirui Wang$^{\spadesuit}$,} \\ 
    {\normalsize \textbf{Wei Wu}$^{\spadesuit}$\textbf{,} \ \ \textbf{Tao Gui}$^{\blacklozenge}$\thanks{{ }{ }Corresponding authors.}\textbf{,} \ \ \textbf{Qi Zhang}$^{\bigstar,\clubsuit}$\textbf{,} \ \ \textbf{Xuanjing Huang}$^{\bigstar}$ } \\
  {$^\bigstar$ \normalsize School of Computer Science, Fudan University, Shanghai, China} \\
  {$^\blacklozenge$ \normalsize Institute of Modern Languages and Linguistics, Fudan University, Shanghai, China} \\
  {$^\clubsuit$ \normalsize Shanghai Collaborative Innovation Center of Intelligent Visual Computing, Fudan University} \\
  {$^\spadesuit$ \normalsize Meituan Inc., Beijing, China} \\
  \texttt{\normalsize \{rzheng20,rbao18,tgui,qz,xjhuang\}@fudan.edu.cn}\\
  \texttt{\normalsize zhouyh21@m.fudan.edu.cn} \\
  }

\begin{document}
\maketitle
\begin{abstract}

Recent works on \emph{Lottery Ticket Hypothesis} have shown that pre-trained language models (PLMs) contain smaller matching subnetworks (winning tickets) which are capable of reaching accuracy comparable to the original models.
However, these tickets are proved to be not robust to adversarial examples, and even worse than their PLM counterparts.
To address this problem, we propose a novel method based on learning binary weight masks to identify robust tickets hidden in the original PLMs.
Since the loss is not differentiable for the binary mask, we assign the hard concrete distribution to the masks and encourage their sparsity using a smoothing approximation of $L_0$ regularization.
Furthermore, we design an adversarial loss objective to guide the search for robust tickets and ensure that the tickets perform well both in accuracy and robustness.
Experimental results show the significant improvement of the proposed method over previous work on adversarial robustness evaluation. 
\end{abstract}

\section{Introduction}
Large-scale pre-trained language models (PLMs), such as BERT \cite{Devlin2019BERTPO}, Roberta \cite{Liu2019RoBERTaAR} and T5 \cite{Raffel2020ExploringTL} have achieved great success in the field of natural language processing.
As more transformer layers are stacked with larger self-attention blocks, the complexity of PLMs increases rapidly.
Due to the over-parametrization of PLMs, some Transformer heads and even layers can be pruned without significant losses in performance  \cite{Michel2019AreSH,Kovaleva2019RevealingTD,Rogers2020API}.

The Lottery Ticket Hypothesis suggests an over-parameterized network contains certain subnetworks (i.e., winning tickets) that can match the performance of the original model when trained in isolation \cite{Frankle2019TheLT}.
\citet{chen2020lottery,  Prasanna2020WhenBP} also find these winning tickets exist in PLMs.
\citet{chen2020lottery} prune BERT in an unstructured fashion and obtain winning tickets at sparsity from 40\% to 90\%.
\citet{Prasanna2020WhenBP} aim at finding structurally sparse tickets for BERT by pruning entire attention heads and MLP.
Previous works mainly focused on using 
winning tickets to reduce model size and speed up training time \cite{Chen2021EarlyBERTEB}, while little work has been done to explore more benefits, such as better adversarial robustness than the original model.

As we all know, PLMs are vulnerable to adversarial examples that are legitimately crafted by imposing imperceptible perturbations on normal examples \cite{Jin2020IsBR,garg2020bae,wang-etal-2021-textflint}.
Recent studies have shown that pruned subnetworks of PLMs are even less robust than their PLM counterparts \cite{Xu2021BeyondPA, Du2021WhatDC}.
\citet{Xu2021BeyondPA} observe that when fine-tuning the pruned model again, the model yields a lower robustness.
\citet{Du2021WhatDC} clarify the above phenomenon further: the compressed models overfit on shortcut samples and thus perform consistently less robust than the uncompressed large model on adversarial test sets. 

In this work, our goal is to find robust PLM tickets that, when fine-tuned on downstream tasks, achieve matching test performance but are more robust than the original PLMs.
In order to make the topology structure of tickets learnable, we assign binary masks to pre-trained weights to determine which connections need to be removed.
To solve discrete optimization problem of binary masks, we assume the masks follow a hard concrete distribution (a soft version of the Bernoulli distribution), which can be solved using Gumbel-Softmax trick \cite{Louizos2018LearningSN}.
We then use an adversarial loss objective to guide the search for robust tickets and an approximate  $L_O$ regularization is used to encourage the sparsity of robust tickets.
Robust tickets can be used as a robust substitute of original PLMs to fine-tune downstream tasks.
Experimental results show that robust tickets achieve a significant improvement in adversarial robustness on various tasks and maintain a matching accuracy.  Our codes are publicly available at \textit{Github}\footnote{https://github.com/ruizheng20/robust\_ticket}.

The main contributions of our work are summarized as follows:
\begin{itemize}
\setlength{\itemindent}{0em}
\setlength{\itemsep}{0em}
\setlength{\topsep}{-0.5em}
    \item We demonstrate that PLMs contain robust tickets with matching accuracy but better robustness than the original network.
    \item We propose a novel and effective technique to find the robust tickets based on learnable binary masks rather than the traditional iterative magnitude-based pruning. 
    \item We provide a new perspective to explain the vulnerability of PLMs on adversarial examples: some weights of PLMs do not contribute to the accuracy but may harm the robustness.
\end{itemize}

\section{Related Work}

\subsection{Textual Adversarial Attack and Defense}
Textual attacks typically generate explicit adversarial examples by replacing the components of sentences with their counterparts and maintaining a high similarity in semantics \cite{ren2019generating} or embedding space \cite{Li2020BERTATTACKAA}.
These  adversarial attackers can be divided into character-level \cite{Gao2018BlackBoxGO}, word-level \cite{ren2019generating, Zang2020WordlevelTA, Jin2020IsBR, Li2020BERTATTACKAA} and multi-level \cite{li2018textbugger}.
In response to adversarial attackers, various adversarial defense methods are proposed to improve model robustness.
Adversarial training solves a min-max robust optimization and is generally considered as one of the strongest defense methods \cite{madry2017towards, zhu_freelb_2020, li2020tavat}.
Adversarial data augmentation (ADA) has been widely adopted to improve robustness by adding textual adversarial examples during training \cite{Jin2020IsBR, Si2021BetterRB}. 
However, ADA is not sufficient to cover the entire perturbed search space, which grows exponentially with the length of the input text.
Some regularization methods, such as smoothness-inducing regularization \cite{Jiang2020SMARTRA} and information bottleneck regularization \cite{wang2020infobert}, are also beneficial for robustness.
Different from the above methods, we dig robust tickets from original BERT, and the subnetworks we find have better robustness after fine-tuning.

\subsection{Lottery Ticket Hypothesis}

Lottery Ticket Hypothesis (LTH) suggests the existence of certain sparse subnetworks (i.e., winning tickets) at initialization that can achieve almost the same test performance compared to the original model \cite{Frankle2019TheLT}.
In the field of NLP, previous works find that the winning tickets also exist in Transformers and LSTM \cite{Yu2020PlayingTL, Renda2020ComparingRA}.
\citet{Evci2020RiggingTL} propose a method to optimize the topology of the sparse network during training without sacrificing accuracy relative to existing dense-to-sparse training methods. 
\citet{chen2020lottery} find that PLMs such as BERT contain winning tickets with a sparsity of 40\% to 90\%, and the winning tickets found in the mask language modeling task can universally be transfered to other downstream tasks.
\citet{Prasanna2020WhenBP} find structurally sparse winning tickets for BERT, and they notice that all subnetworks (winning tickets and randomly pruned subnetworks) have comparable performance when fine-tuned on downstream tasks.
\citet{Chen2021EarlyBERTEB} propose an efficient BERT training method using Early-bird lottery tickets to reduce the training time and inference time.
Some recent studies have tried to dig out more features of winning tickets.
\citet{Zhang2021CanSS} demonstrate that even in biased models (which focus on spurious correlations) there still exist unbiased winning tickets.
\citet{Liang2021SuperTI} observe that at a certain sparsity, the generalization performance of the winning tickets can not only match but also exceed that of the full model. 
\cite{Du2021WhatDC, Xu2021BeyondPA} show that the winning tickets that only consider accuracy are over-fitting on easy samples and generalize poorly on adversarial examples. 
Our work makes the first attempt to find the robust winning tickets for PLMs.

\subsection{Robustness in Model Pruning}

Learning to identify a subnetwork with high adversarial robustness is widely discussed in the field of computer vision.
Post-train pruning approaches require a pre-trained model with adversarial robustness before pruning \cite{Sehwag2019TowardsCA, Gui2019ModelCW}.
In-train pruning methods integrate the pruning process into the robust learning process, which jointly optimize the model parameters and pruning connections \cite{Vemparala2021AdversarialRM, Ye2019AdversarialRV}. 
\citet{Sehwag2020HYDRAPA} integrate the robust training objective into the pruning process and remove the connections based on importance scores.
In our work, we focus on finding robust tickets hidden in original PLMs rather than pruning subnetworks from a robust model.

\section{The Robust Ticket Framework}
In this section, we propose a novel pruning method to extract robust tickets of PLMs by learning binary weights masks with an adversarial loss objective.
Furthermore, we articulate the Robust Lottery Ticket Hypothesis: the full PLM contains subnetworks (robust tickets) that can achieve better adversarial robustness and comparable accuracy.

\subsection{Revisiting Lottery Ticket Hypothesis}
Denote $f(\theta)$ as a PLM with parameters $\theta$ that has been fine-tuned on a downstream task. 
A subnetwork of $f(\theta)$ can be denoted as $f(m \odot \theta)$, where $m$ are binary masks with the same dimension as $\theta$ and $\odot$ is the Hadamard product operator.
LTH suggests that, for a network initialized with $\theta_0$, the Iterative Magnitude Pruning (IMP) can identify a mask $m$, such that the subnetwork $f(x;m \odot \theta_0)$ can be trained to almost the same performance to the full model $f(\theta_0)$ in a comparable number of iterations.
Such a subnetwork $f(x;m \odot \theta_0)$ is called as \emph{winning tickets}, including both the structure mask $m$ and initialization $\theta_0$. 
IMP iteratively removes the weights with the smallest magnitudes from $m \odot \theta$ until a certain sparsity is reached.
However, the magnitude-based pruning is not suitable for robustness-aware techniques \cite{Vemparala2021AdversarialRM, Sehwag2020HYDRAPA}.

\subsection{Discovering Robust Tickets}

Our goal is to learn the sparse subnetwork, however, the training loss is not differentiable for the binary masks.
A simple choice is to adopt a straight-through estimator to approximate the derivative \cite{Bengio2013EstimatingOP}. 
Unfortunately, this approach ignores the Heaviside function in the likelihood and results in biased gradients. Thus, we resort to a practical method to learn sparse neural networks \cite{Louizos2018LearningSN}.

In our method, we assume each mask $m_i$ to be a independent random variable that follows a hard concrete distribution $\mathrm{HardConcrete}(\log \alpha_i, \beta_i)$ with temperature $\beta_i$ and location $\alpha_i$ \cite{Louizos2018LearningSN}:
\begin{flalign}
& \mu_i  \sim \mathcal{U}\left(0,1\right),\\
 & s_i = \sigma\left(\frac{1}{\beta_i}\left(\log\frac{\mu_i}{1-\mu_i}+\log \alpha_i\right)\right) \label{gumbel}, \\
& m_i  = \min\left(1,\max\left(0,s_i\left(\zeta-\gamma\right)+\gamma\right)\right), \label{stretch}
\end{flalign}
where $\sigma$ denotes the sigmoid function, $\gamma=-0.1$, $\zeta=1.1$ are constants, and $u_i$ is the sample drawn from uniform distribution $\mathcal{U}(0,1)$.
The random variable $s_i$ follows a binary concrete (or Gumbel-Softmax) distribution, which is a smoothing approximation of the discrete Bernoulli distribution \cite{Maddison2017TheCD, Jang2017CategoricalRW}. 
Samples from the binary concrete distribution are identical to samples from a Bernoulli distribution with probability $\alpha_i$ as $\beta_i\rightarrow 0$. 
The location $\alpha_i$ in (\ref{gumbel}) allows for gradient-based optimization through reparametrization tricks.
Using (\ref{stretch}), the $s_i$ larger than $\frac{1-\gamma}{\zeta-\gamma}$ is rounded to $1$, whereas the value smaller than $\frac{-\gamma}{\zeta-\gamma}$ is rounded to $0$. 
To encourage the sparsity, we penalize the $L_0$ complexity of masks based on the probability which are non-zero:
\begin{equation}
    \mathcal R(m)=\frac{1}{\left|m\right| }\sum_{i=1}^{\left|m\right| } \sigma\left(\log \alpha_i-\beta_i \log \frac{-\gamma}{\zeta} \right).
\end{equation}
During the inference stage, the mask $\hat{m}_i$ can be estimated through a hard concrete gate:
\begin{equation}
      \min\left(1,\max\left(0,\sigma\left(\log \alpha_i\right)\left(\zeta-\gamma\right)+\gamma\right)\right).
\end{equation}

\subsubsection{Adversarial Loss Objective}
To find the connections responsible for adversarial robustness, we incorporate the adversarial loss into the mask learning objective:
\begin{equation}
\min_{m} \underbrace{\mathbb{E}_{(x,y)\sim\mathcal{D}} \max_{\lVert \delta \rVert  \leq \epsilon} \mathcal{L}\left(f(x+\delta; m\odot \theta),y\right)}_{\mathcal{L}_{adv}(m)},\label{inner} 
\end{equation}
where $(x,y)$ is a data point from dataset $\mathcal{D}$, $\delta$ is the perturbation that constrained within the $\epsilon$ ball.
The inner maximization problem in (\ref{inner})
is to find the worst-case adversarial examples to maximize the classification loss, while the outer minimization problem in (\ref{inner}) aims at optimizing the masks to minimize the loss of adversarial examples, i.e., $\mathcal{L}_{adv}(m)$.

Adversarial attack method, typically with PGD, can be used to solve the inner maximization problem.
PGD applies the $K$-step stochastic gradient descent to search for the perturbation $\delta$ \cite{madry2017towards}:
\begin{equation}
\delta_{k+1} = \prod_{\lVert \delta \rVert \leq \epsilon}\left(\delta_{k}+
\eta \frac{g\left(\delta_{k}\right)}{\rVert g\left(\delta_{k}\right)\rVert}\right),
\end{equation}
where $g\left(\delta_{k}\right)=\nabla_x\mathcal{L}\left(f(x+\delta_{k}; m\odot \theta),y\right)$, $\delta_{k}$ is the perturbation in $k$-th step and $\prod_{\lVert \delta \rVert \leq \epsilon}(\cdot)$ projects the perturbation back onto the Frobenius normalization ball. Then robust training optimizes the network on adversarially perturbed input $x+\delta_K$.
Through the above process, we can conveniently obtain a large number of adversarial examples for training.

By integrating the $L_0$ complexity regularizer into the training process of masks, our adversarial loss objective becomes:
\begin{equation}
    \min_{m} \mathcal{L}_{adv}(m) + \lambda \mathcal R(m), \label{adv_object}
\end{equation}
where $\lambda$ denotes regularization strength. 

\subsubsection{Effect of Regularization Strength} \label{effect_lambda}
\begin{figure}[t]
\centering
\includegraphics[height=4.5cm]{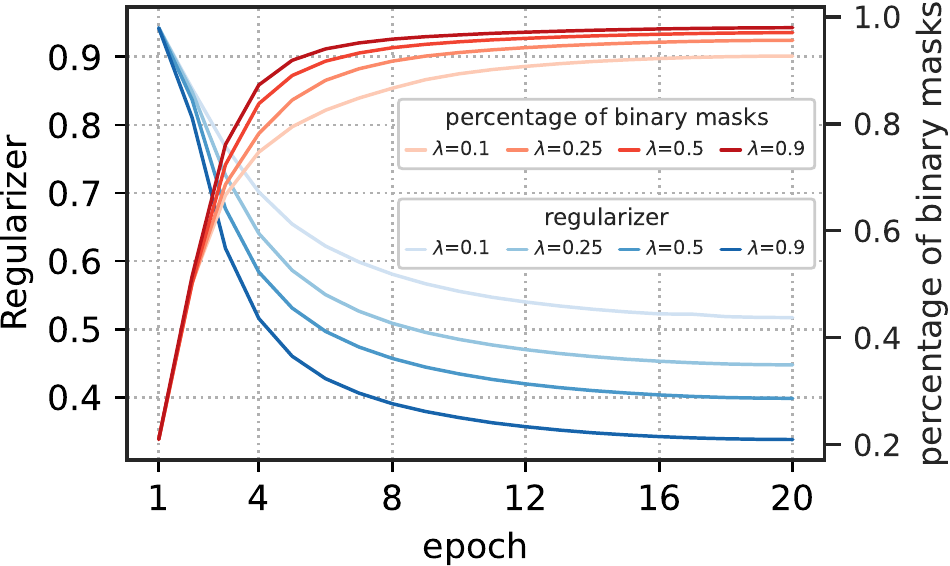}
\caption{Effect of regularization strength $\lambda$ on regularizer $\mathcal R(m)$, and the percentage of masks that exact $0$ and $1$.}
\label{fig:sst2_regular}
\end{figure}

The selection of the regularization strength $\lambda$ decides the quality of robust tickets. Results carried on SST-2 in Fig.\ref{fig:sst2_regular} show that eventually more than $90\%$ of the masks will be very close to $0$ or $1$, and the $L_0$ complexity regularizer $\mathcal R(m)$ will converge to a fixed value.
As $\lambda$ increases, $\mathcal R(m)$ decreases (the sparsity of the subnetwork increases).
The training of the adversarial loss objective in (\ref{adv_object}) is insensitive to the $\lambda$, and in all experiments, $\lambda$ is chosen in the range $[0.1, 1]$. In the Appendix \ref{appendix0}, we show more details about mask learning process.

\subsection{Drawing and Retraining Winning Tickets}

After training the masks $m$, we use the location parameters $\log \alpha$ of masks to extract robust tickets.
For the Gumbel-Softmax distribution in (\ref{gumbel}), $\alpha_i$ is the expectation (confidence) of random variable $s_i$, i.e, $\mathbb{E}\{s_i\}=\alpha_i$.
Thus, we prune the weights whose masks have the smallest expectation.
We prune all attention heads and intermediate neurons in an unstructured manner, which empirically has better performance than structured pruning.
Unlike the Lottery Ticket Hypothesis that requires iterative magnitude pruning, the proposed method is a one-shot pruning method that can obtain subnetworks of any sparsity.
Then we retrain (i.e., fine-tune) the robust tickets $f(m \odot \theta_0)$ on downstream tasks.

\subsection{Robust Lottery Tickets Hypothesis}

In the context of adversarial robustness, we seek winning tickets that balance accuracy and robustness, and then we state and demonstrate Robust Lottery Tickets Hypothesis.

\paragraph{Robust Lottery Tickets Hypothesis:}
A pre-trained language model, such as BERT, contains some subnetworks (robust tickets) initialized by pre-trained weights, and when these subnetworks are trained in isolation, they can achieve better adversarial robustness and comparable accuracy. 
In addition, robust tickets retain an important characteristic of traditional lottery tickets  \textemdash  the ability to speed up the training process.

The practical merits of Robust Lottery Ticket Hypothesis: 
1)	It provides an effective pruning method that can reduce memory constraints during inference time by identifying well-performing smaller networks which can fit in memory.
2)	Our proposed robust ticket is more robust than the existing defense methods, so it can be used as a defense method.

\section{Experiments}
We conduct several experiments to demonstrate the effectiveness of our method.
We first compare the proposed method with baseline methods in terms of clean accuracy and robust evaluation.
Then, we perform an ablation study to illustrate the role of sparse mask learning and adversarial loss objective in our method.
In addition, we try to further flesh out our method with several additional analysis experiments.
Following the official BERT implementation \cite{Devlin2019BERTPO, wolf2020huggingfaces},
we use BERT$_{\mathrm{BASE}}$ as our backbone model for all experiments.

\subsection{Datasets}
We evaluate our method mainly on three text classification datasets: 
Internet Movie Database \citep[IMDB,][]{maas-etal-2011-learning}
, AG News corpus \citep[AGNEWS,][]{zhang2016characterlevel} and Stanford Sentiment Treebank of binary classification \citep[SST-2,][]{socher2013recursive}.
We also test our method on other types of tasks in GLUE, such as MNLI, QNLI, QQP.
The labels of GLUE test sets are not available, so GLUE test sets cannot be used for adversarial attacks. The results of GLUE tasks are tested on the official development set, and we divide $10\%$ training data as the development set.

\subsection{Baselines}
We compare our \textbf{RobustT} (\textbf{Robust T}ickets) with recently proposed adversarial defense methods and the standard lottery ticket.

\textbf{Fine-tune} \cite{Devlin2019BERTPO}: The official BERT implementation on downstream tasks.
\textbf{FreeLB} \cite{zhu_freelb_2020}: An enhanced gradient-based adversarial training method which is not targeted at specific attack methods.
\textbf{InfoBERT} \cite{wang2020infobert}: A learning framework for robust model fine-tuning from an information-theoretic perspective. This method claims that it has obtained a better representation of data features.
\textbf{LTH} \cite{chen2020lottery}: For a range of downstream tasks, BERT contains winning lottery tickets at 40\% to 90\% sparsity.
\textbf{Random}: Subnetworks with the same layer-wise sparsity of the above RobustT, but their structures are randomly pruned from the original BERT.

\subsection{Robust Evaluation}

Three widely accepted attack methods are used to verify the ability of our proposed method against baselines \cite{Li2021SearchingFA}.
\textbf{BERT-Attack}
\cite{Li2020BERTATTACKAA} is a method using BERT to generate adversarial text, and thus the generated adversarial examples are fluent and semantically preserved.
\textbf{TextFooler}
\cite{Jin2020IsBR} first identify the important words in the sentences, and then replace them with synonyms that are semantically similar and grammatically correct until the prediction changes.
\textbf{TextBugger}
\cite{li2018textbugger} is an adversarial attack method that generates misspelled words by using character-level and word-level perturbations. 

The evaluation metrics adopted in our experimental analyses are listed as follows:
\textbf{Clean accuracy (Clean$\%$)} denotes the accuracy on the clean test dataset.
\textbf{Accuracy under attack (Aua$\%$)} refers to the model’s prediction accuracy facing specific adversarial attacks.
\textbf{Attack success rate (Suc$\%$)} is the ratio of the number of texts successfully perturbed by an attack method to the total number of texts to be attempted.
\textbf{Number of Queries (\#Query)} is the average number of times the attacker queries the model, which means the more the average query number is, the harder the defense model is to be compromised.
For a robust method, higher clean accuracy, accuracy under attack, and query times are expected, as well as lower attack success rate.

\subsection{Implementation Details}

We fine-tune the original BERT using the default settings on downstream tasks. We train 20 epochs to discover the robust tickets from the fine-tuned BERT, and then we retrain the robust tickets using default settings of BERT-base.
The $K$-step PGD requires $K$ forward-backward passes through the network, which is time consuming.
Thus, we turn to FreeLB, which accumulates gradients in multiple forward passes and then passing gradients backward once.
For our approach, we prune robust tickets in the range of 10\% and 90\% sparsity and report the best one in terms of robustness in our main experiments.
For a fair comparison, the sparsity of LTH is the same as that of robust tickets.
All experimental results are the average of $5$ trials with different seeds.
More implementation details and hyperparameters are provided in the Appendix \ref{appendix1}.
We implement all models in MindSpore.

\subsection{Main Results on Robustness Evaluation}

\begin{table*}[ht]
\renewcommand\arraystretch{1.2}
\setlength\tabcolsep{5pt}
\centering
\small
\begin{tabular}{c|l|c|ccc|ccc|ccc}
\hline
\hline
\multicolumn{1}{c|}{\multirow{2}{*}{\textbf{Dataset}}} &
\multicolumn{1}{c|}{\multirow{2}{*}{\textbf{Method}}} &
\multicolumn{1}{c|}{\multirow{2}{*}{\textbf{Clean$\%$}}} &
\multicolumn{3}{c|}{\textbf{BERT-Attack}} & 
\multicolumn{3}{c|}{\textbf{TextFooler}} & 
\multicolumn{3}{c}{\textbf{TextBugger}} \\ \cline{4-12}
\multicolumn{1}{c|}{} & \multicolumn{1}{c|}{} & \multicolumn{1}{c|}{}& 
\multicolumn{1}{c}{\textbf{Aua$\%$}} &
\multicolumn{1}{c}{\textbf{ Suc$\%$}} & 
\multicolumn{1}{c|}{\textbf{\#Query} } & 
\multicolumn{1}{c}{\textbf{Aua$\%$}} & 
\multicolumn{1}{c}{\textbf{Suc$\%$} } &  
\multicolumn{1}{c|}{\textbf{\#Query} } & 
\multicolumn{1}{c}{\textbf{Aua$\%$} } & 
\multicolumn{1}{c}{\textbf{Suc$\%$} } & 
\multicolumn{1}{c}{\textbf{\#Query} } \\ \cline{1-12}
\hline
\multirow{7}{*}{\textbf{IMDB}}
& Fine-tune & $94.1$ & $7.8$ & $91.7$ & $1572.2$ & $12.2$ & $87.0$ & $1209.8$ & $25.8$ & $72.5$ & $783.2$ \\ 
&$\textrm{LTH}_{20\%}$  & $94.0$ & $3.6$ & $96.2$ & $1074.44$ & $7.2$ & $92.3$ & $894.1$ & $16.0$ & $83.0$ & $574.0$ \\
&FreeLB & $94.8$ & $22.6$ & $76.2$ & $1954.7$ & $27.2$ & $71.3$ & $1479.1$ & $36.0$ & $62.0$ & $907.3$ \\
&InfoBERT  & $\mathbf{95.2}$ & $26.0$ & $72.7$ & $2326.0$ & $32.4$ & $66.0$ & $1572.2$ & $43.6$ & $54.2$ & $969.8$ \\
% \cline{2-12}
&$\textrm{Rand}_{20\%}$ & $93.1$ & $6.8$ & $92.8$ & $731.5$ & $7.4$ & $92.1$ & $598.7$  & $8.4$ & $91.9$ & $464.3$ \\
&$\textbf{RobustT}_{20\%}$ & $93.8$ & $\mathbf{55.2}$ & $\mathbf{41.2}$ & $\mathbf{3128.0}$ & $\mathbf{55.6}$ & $\mathbf{40.7}$ & $\mathbf{1988.4}$  & $\mathbf{57.6}$ & $\mathbf{38.6}$ & $\mathbf{1149.1}$ \\
\hline
\multirow{7}{*}{\textbf{AGNEWS}}&
Fine-tune & $94.7$ & $3.8$ & $96.0$ & $436.7$ & $14.9$ & $84.2$ & $333.2$ & $41.5$ & $56.1$ & $178.3$ \\ 
&$\textrm{LTH}_{40\%}$  & $93.7$ & $2.5$ & $97.3$ & $394.4$ & $11.0$ & $88.3$ & $295.2$ & $36.8$ & $60.7$ & $179.7$ \\
&FreeLB  & $\mathbf{95.2}$ & $10.8$ & $88.6$ & $563.9$ & $24.3$ & $74.4$ & $394.6$ & $51.7$ & $45.5$ & $190.4$ \\
&InfoBERT  & $94.4$ & $11.1$ & $88.3$ & $517.0$ & $25.1$ & $73.4$ & $374.7$ & $47.9$ & $49.3$ & $193.1$ \\
% \cline{2-12}
&$\textrm{Rand}_{40\%}$ & $94.0$ & $1.3$ & $98.6$ & $357.2$ & $6.3$ & $93.2$ & $275.1$  & $27.5$ & $70.1$ & $148.7$ \\
&$\textbf{RobustT}_{40\%}$ & $94.9$ & $\mathbf{12.1}$ & $\mathbf{87.2}$ & $\mathbf{607.7}$ & $\mathbf{28.5}$ & $\mathbf{70.0}$ & $\mathbf{442.1}$  & $\mathbf{53.4}$ & $\mathbf{43.7}$ & $\mathbf{207.8}$ \\
\hline
\multirow{7}{*}{\textbf{SST-2}}&
Fine-tune & $92.0$ & $2.9$ & $96.8$ & $114.2$ & $5.0$ & $94.6$ & $98.4$ & $29.4$ & $68.3$ & $49.7$ \\ 
&$\textrm{LTH}_{30\%}$ & $92.1$ & $2.2$ & $97.6$ & $98.9$ & $4.1$ & $95.5$ & $90.5$ & $29.1$ & $68.4$ & $49.6$ \\
&FreeLB  & $91.6$ & $10.2$ & $88.9$ & $154.6$ & $14.4$ & $84.2$ & $123.8$ & $\mathbf{42.4}$ & $\mathbf{53.7}$ & $\mathbf{54.9}$ \\
&InfoBERT  & $\mathbf{92.1}$ & $14.4$ & $84.4$ & $162.3$ & $18.3$ & $80.1$ & $121.4$ & $40.3$ & $56.3$ & $51.2$ \\
% \cline{2-12}
&$\textrm{Rand}_{30\%}$& $83.2$ & $2.1$ & $97.5$ & $89.4$ & $2.4$ & $97.1$ & $75.6$  & $16.5$ & $80.2$ & $44.2$ \\
&$\textbf{RobustT}_{30\%}$ & $90.9$ & $\mathbf{17.9}$ & $\mathbf{80.3}$ & $\mathbf{164.9}$ & $\mathbf{26.7}$ & $\mathbf{70.6}$ & $\mathbf{149.8}$  & $42.1$ & $53.7$ & $53.9$ \\

\hline
\hline
\end{tabular}
\caption{Main results on adversarial robustness evaluation. Fine-tuning \textbf{RobustT} for downstream tasks achieves a significant improvement of robustness. The percentage on the subscript denotes the sparsity of the subnetworks. 
The best performance is marked in bold. \textbf{Suc$\%$} lower is better.}
\label{tab:results_main}
\end{table*}

\begin{table}[t]
\renewcommand\arraystretch{1.2}
\setlength\tabcolsep{5pt}
\centering
\small
\begin{tabular}{c|l|c|c|c}
\hline
\hline
\multicolumn{1}{c|}{\multirow{2}{*}{\textbf{Dataset}}} &
\multicolumn{1}{c|}{\multirow{2}{*}{\textbf{Method}}} &
\multicolumn{1}{c|}{\multirow{2}{*}{\textbf{Clean$\%$}}} &
\multicolumn{2}{c}{\textbf{Aua$\%$}}
\\ \cline{4-5}
\multicolumn{1}{c|}{} & \multicolumn{1}{c|}{} & \multicolumn{1}{c|}{}& 
\multicolumn{1}{c|}{\textbf{\scriptsize TextFooler}} &
\multicolumn{1}{c}{\textbf{\scriptsize TextBugger}}   \\ 
\cline{1-4}
\hline
\multirow{4}{*}{\textbf{QNLI}}&
Fine-tune & $\mathbf{91.6}$ & $4.7$  & $10.5$ \\ 
&FreeLB  & $90.5$ & $12.8$ & $12.0$ \\
&InfoBERT  & $91.5$ & $16.4$ & $20.9$ \\
% \cline{2-5}
& $\textbf{RobustT}_{30\%}$ & $91.5$ & $\mathbf{17.0}$ & $\mathbf{25.9}$  \\
\hline
\multirow{4}{*}{\textbf{MNLI}}&
Fine-tune & $\mathbf{84.4}$ & $7.7$  & $4.3$   \\ 
&FreeLB  & $82.9$ & $11.0$ & $8.4$  \\
&InfoBERT  & $84.1$ & $10.8$ & $8.4$  \\
&$\textbf{RobustT}_{30\%}$ & $84.0$ & $\mathbf{18.4}$ & $\mathbf{22.6}$   \\
\hline
\multirow{4}{*}{\textbf{QQP}}&
Fine-tune & $91.3$ & $24.8$  & $27.8$ \\ 
&FreeLB & $91.2$ & $27.4$  & $28.1$  \\
&InfoBERT & $\mathbf{91.9}$ & $34.4$  & $35.9$  \\
&$\textbf{RobustT}_{30\%}$ & $91.5$ & $\mathbf{47.2}$  & $\mathbf{46.0}$ \\
\hline
\hline
\end{tabular}
\caption{Adversarial robustness evaluation of RobustT on QNLI, MNLI and QQP datasets. Compare with the original BERT, fine-tuning on robust tickets improves the adversarial
robustness.}
\label{tab:glue}
\end{table}

\begin{table}[t]
\renewcommand\arraystretch{1.4}
\setlength\tabcolsep{5pt}
\centering
\small
\begin{tabular}{l|lcc}
\hline
\hline
\textbf{Dataset}   & \multicolumn{1}{c}{\textbf{Method}} & \textbf{Clean$\%$} & \textbf{Aua$\%$} \\ \hline
 \multirow{3}{*}{$\textbf{IMDB}$}   & $\textbf{RobustT}_{20\%}$               &$93.8$ & $\mathbf{55.6}$   \\
 & \quad \textbf{w/o} Mask Leaning   & $\mathbf{94.0}$  & $15.1$   \\ 
 & \quad \textbf{w/o} Adv                   & $93.4$     & $5.4$   \\ \hline
\multirow{3}{*}{$\textbf{AGNEWS}$} & $\textbf{RobustT}_{40\%} $             & $\mathbf{94.9}$     & $\mathbf{28.5}$   \\ 
& \quad \textbf{w/o} Mask Learning                      &  $94.2$    & $16.1$   \\ 
& \quad \textbf{w/o} Adv                  & $94.5$     & $8.8$   \\ \hline
\multirow{3}{*}{$\textbf{SST-2}$}   & $\textbf{RobustT}_{30\%}$               & $90.9$     & $\mathbf{26.7}$   \\ 
& \quad \textbf{w/o} Mask Learning & $\mathbf{92.2}$ & $6.2$   \\ 
& \quad \textbf{w/o} Adv   & $91.2$  & $3.5$ \\ \hline\hline

\end{tabular}
\caption{Ablation study on text classification datasets. \textbf{Aua$\%$} is obtained after using TextFooler attack.}
\label{tab:ablation_main}
\end{table}

Table \ref{tab:results_main} shows the results of robust tickets and other baselines under adversarial attack. 
We can observe that: 1) Original BERT and BERT-tickets fail to perform well on adversarial robustness evaluation, and the BERT-tickets even show lower robustness than BERT, indicating that it is difficult for the pruned subnetworks to fight against adversarial attacks when only test accuracy is considered. 
This result is consistent with the results in \cite{Du2021WhatDC, Xu2021BeyondPA}.
2) The proposed robust ticket achieves a significant improvement of robustness over the original BERT and other adversarial defense methods. 
Robust tickets use a better robust structure to resist adversarial attacks, which is different from the previous methods aimed at solving robust optimization problems.
3) In both AGNEWS and IMDB, the randomly pruned subnetwork loses only about $1$ performance point in test accuracy, but performs poorly in adversarial robustness. 
This suggests that robust tickets are more difficult to discovered than traditional lottery tickets.
4) Robust tickets sacrifice accuracy performance in SST-2 and IMDB. We speculate that this may be due to the trade-off between accuracy and robustness \cite{Tsipras2019RobustnessMB}.

We also evaluate the performance of our proposed method on more tasks.
From Table \ref{tab:glue}, we can see that our proposed method yields significant improvements of robustness over the original BERT on QNLI, MNLI and QQP datasets.
There is a significant improvement even compared with InfoBERT and FreeLB.

\subsection{Ablation Study}
\begin{figure*}[t]
	\centering
	\subfigure[IMDB]{
		\centering
	\begin{minipage}[t]{0.31\textwidth}
	\includegraphics[width=5.1cm]{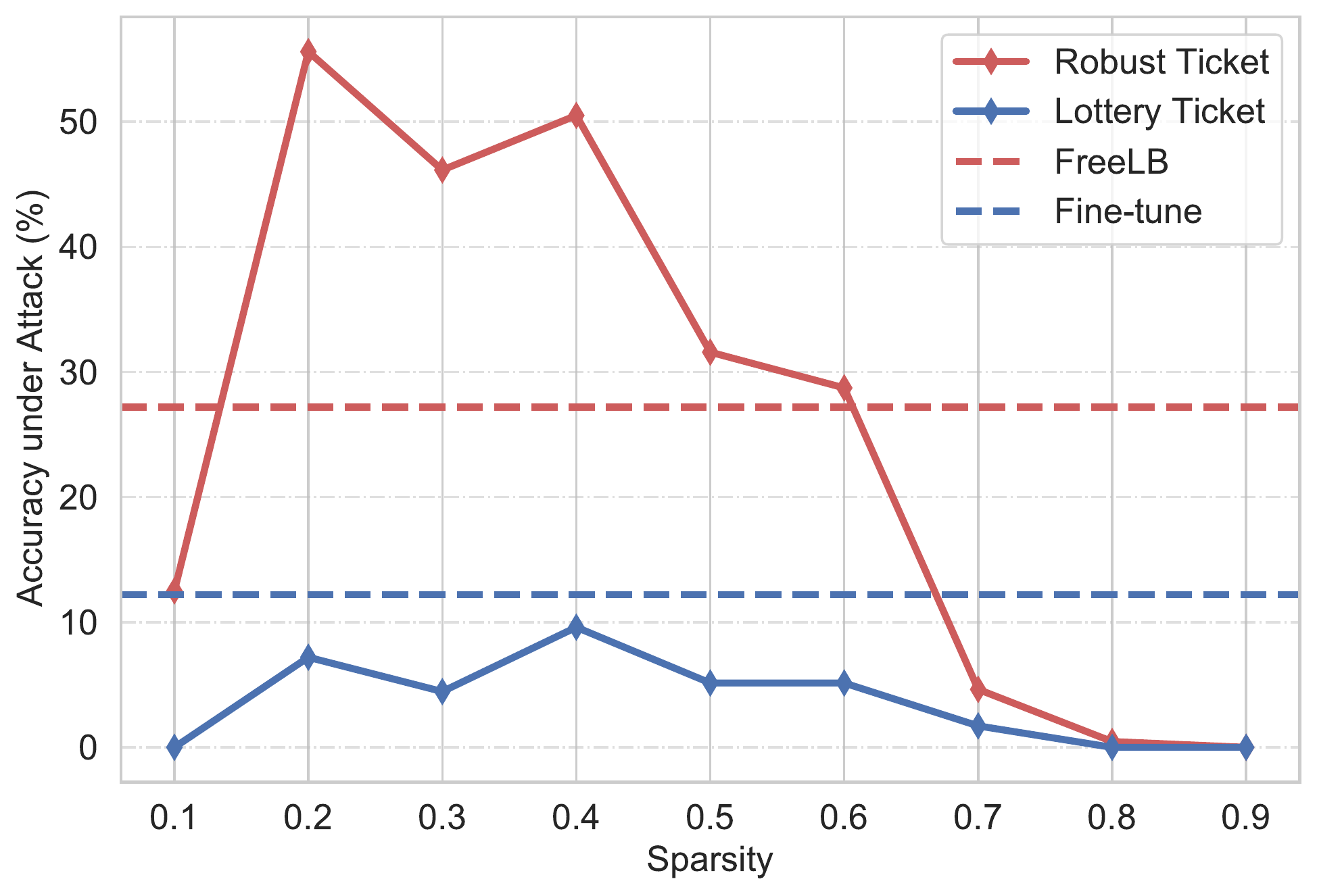}
	\includegraphics[width=5.1cm]{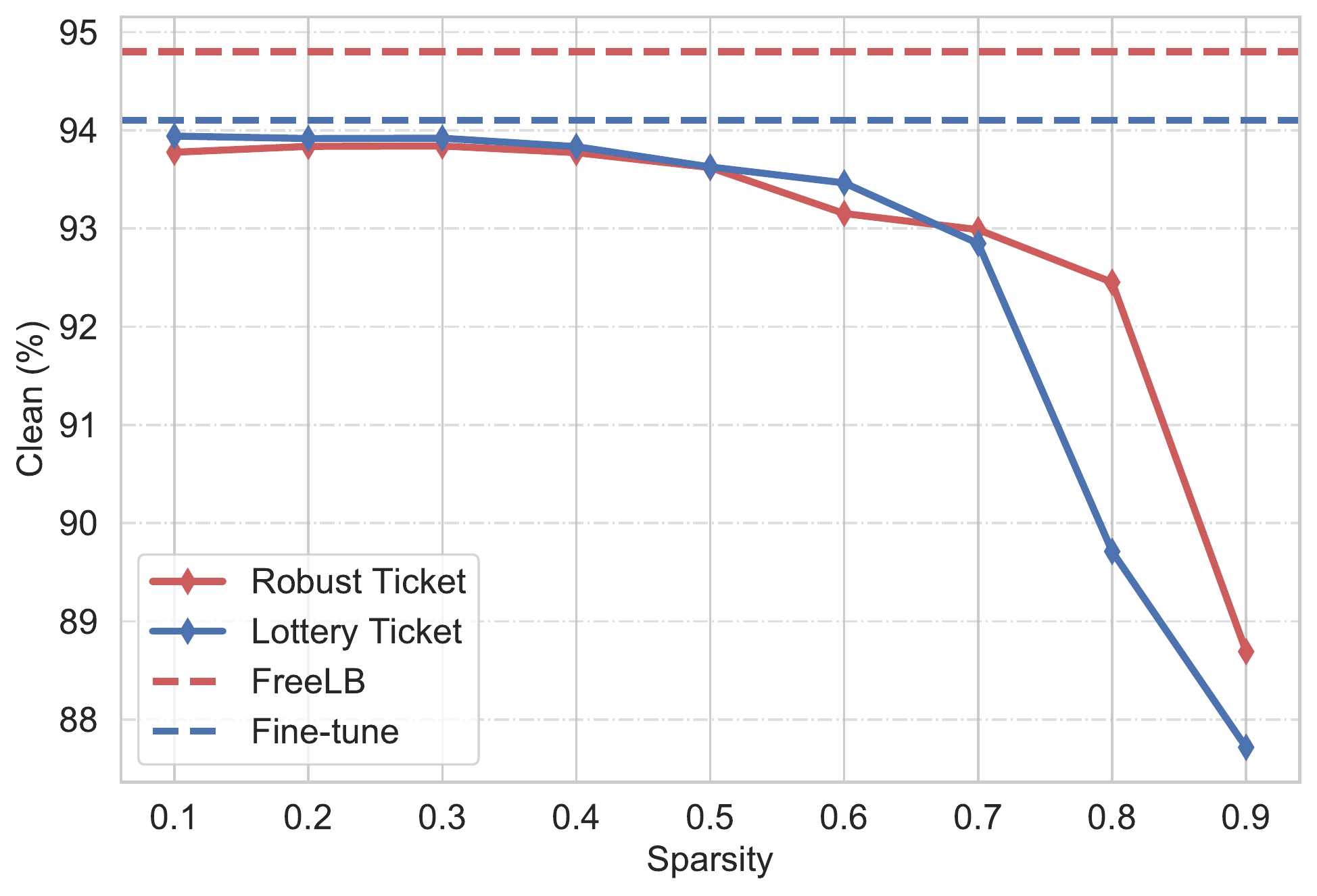}
	\end{minipage}
	}
	\subfigure[AGNEWS]{
		\centering
	\begin{minipage}[t]{0.31\textwidth}
		\includegraphics[width=5.1cm]{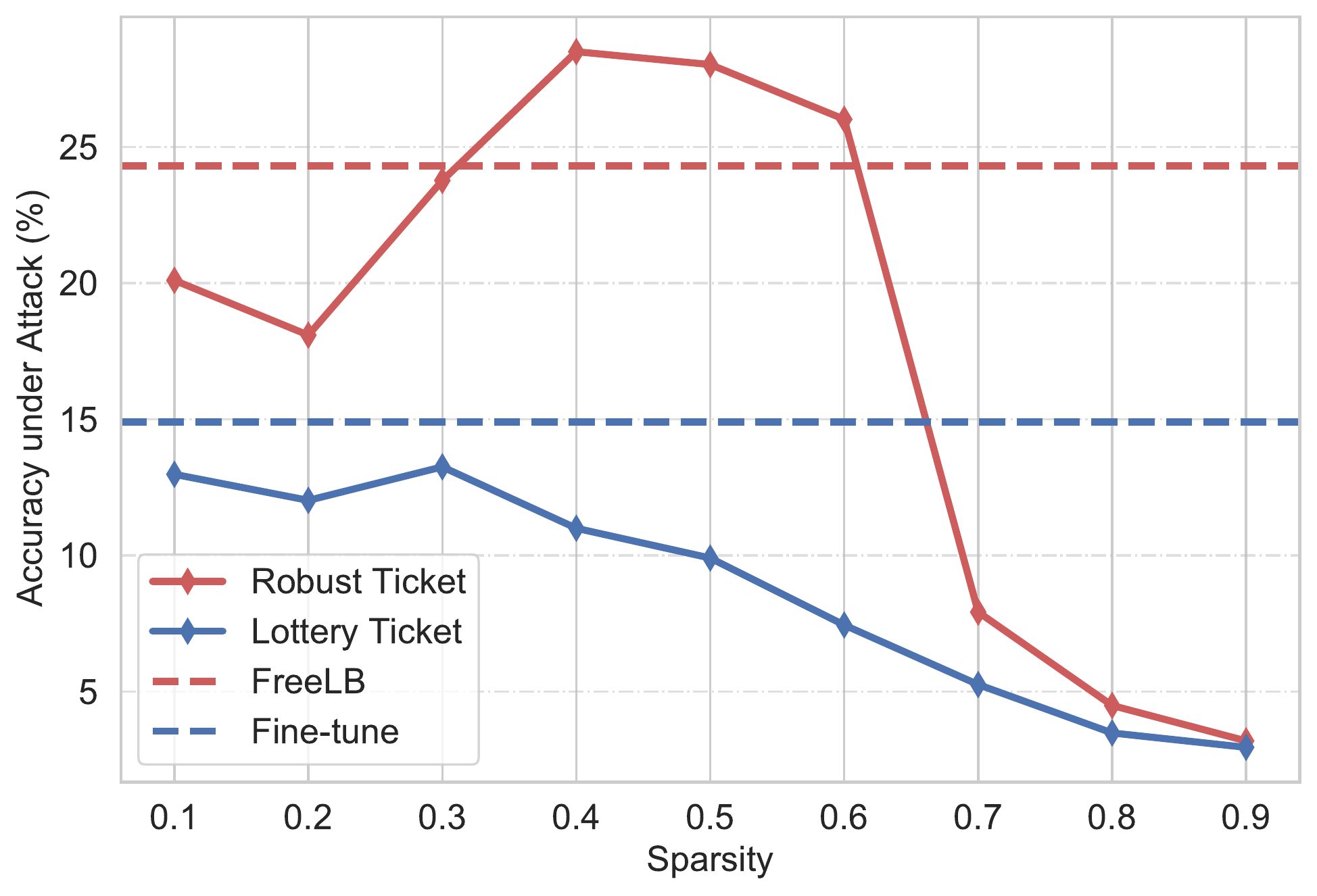}
		\includegraphics[width=5.1cm]{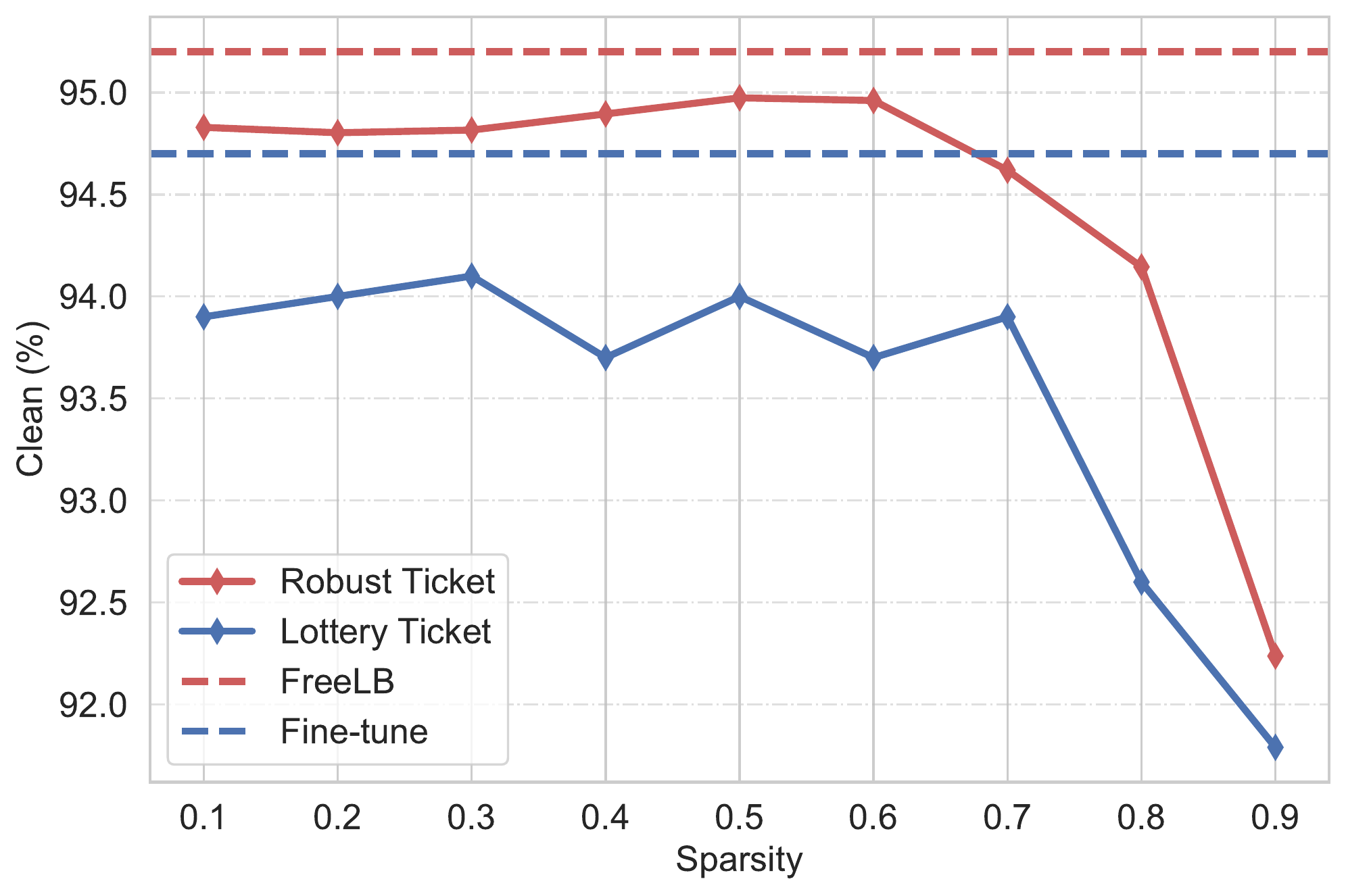}
	\end{minipage}
	}
	\subfigure[SST-2]{
		\centering
	\begin{minipage}[t]{0.31\textwidth}
		\includegraphics[width=5.1cm]{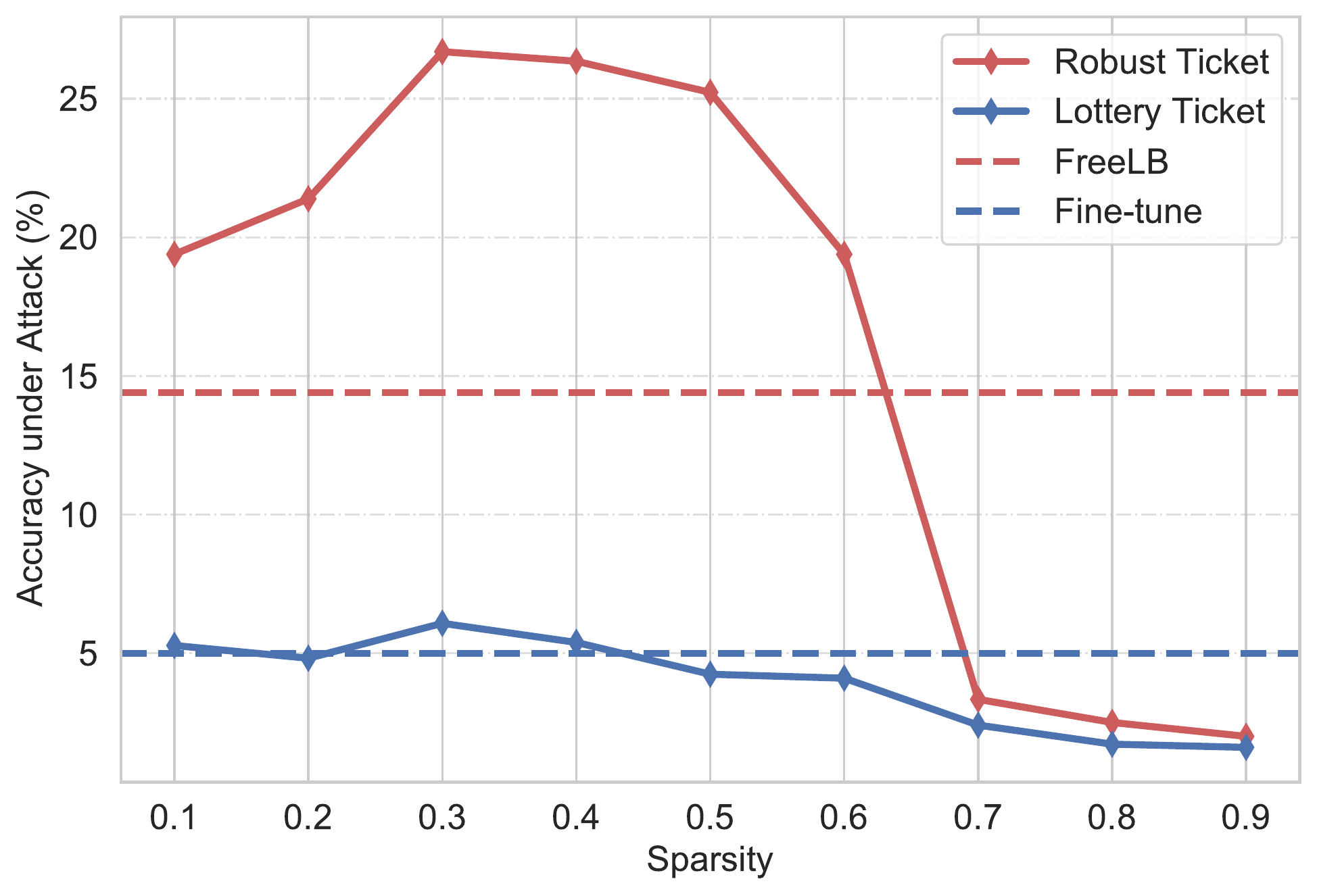}  	
		\includegraphics[width=5.1cm]{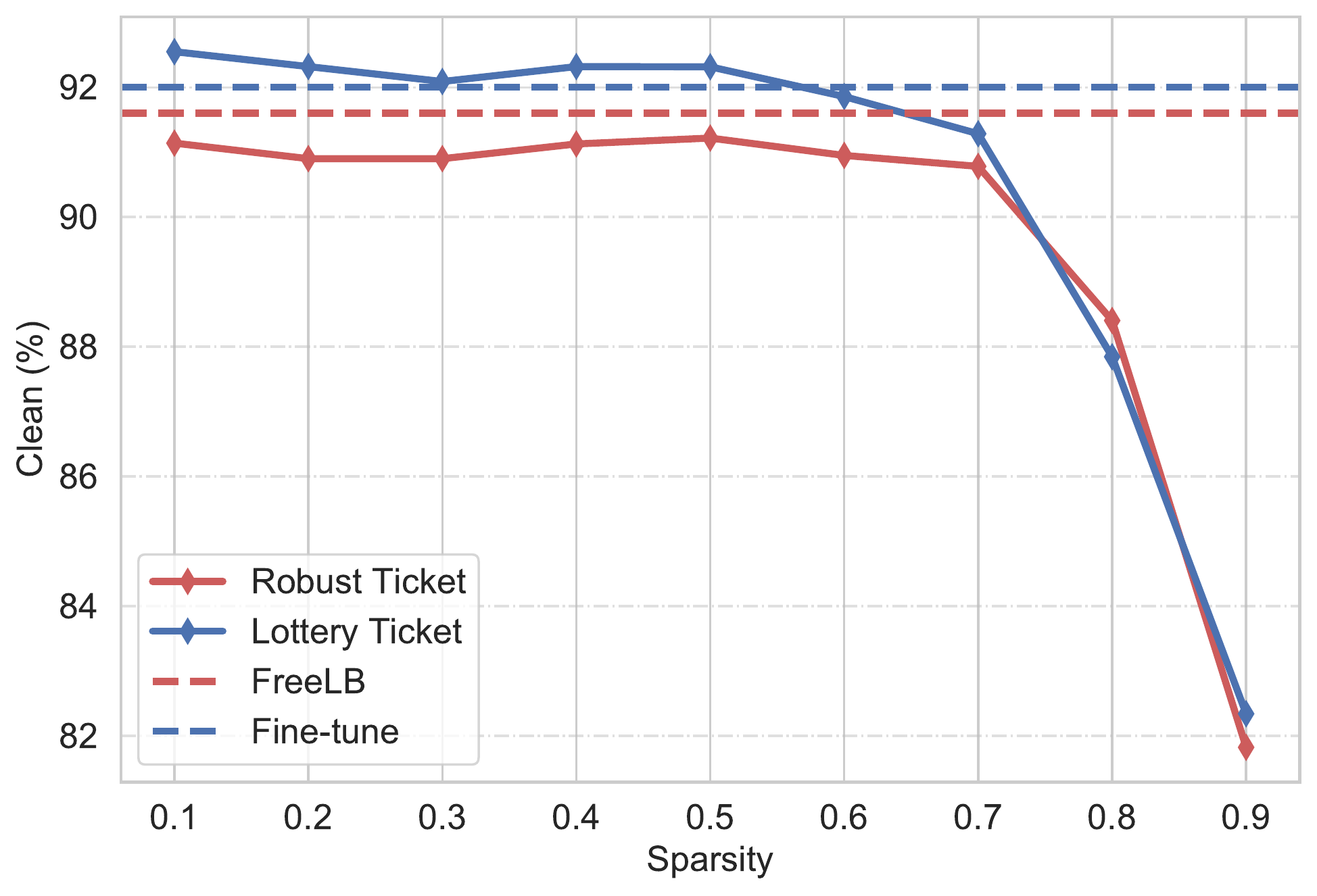}
	\end{minipage}
	}
\caption{Fine-tuning evaluation results of the robust ticket, the traditional lottery ticket, FreeLB and the original BERT fine-tuning under various sparsity levels. The adversarial robustness improves as the compression ratio grows until a certain threshold, then the robustness deteriorates. \textbf{Aua$\%$} is obtained after using TextFooler attack.}
\label{fig:sparse}
\end{figure*}

To better illustrate the contribution of each component of our method, we perform the ablation study by removing the following components: sparse mask learning (but with IMP instead) and adversarial loss objective (Adv).
The test results are shown in Table \ref{tab:ablation_main}.
We can observe that: 1) Mask learning is important for performance and IMP does not identify robust subnetworks well \cite{Vemparala2021AdversarialRM}. 2) Without adversarial loss objective, the proposed method identifies subnetworks that perform well in terms of clean accuracy, but does not provide any improvement in terms of robustness.

\section{Discussion}
In this section, we study how the implementation of robust tickets affects the model’s robustness.

\subsection{Impact of Sparsity on Robust Tickets}

\begin{figure*}[t]
\centering
\subfigure[IMDB]{
\label{fig:heatmap_mnli}
\includegraphics[width=5.1cm]{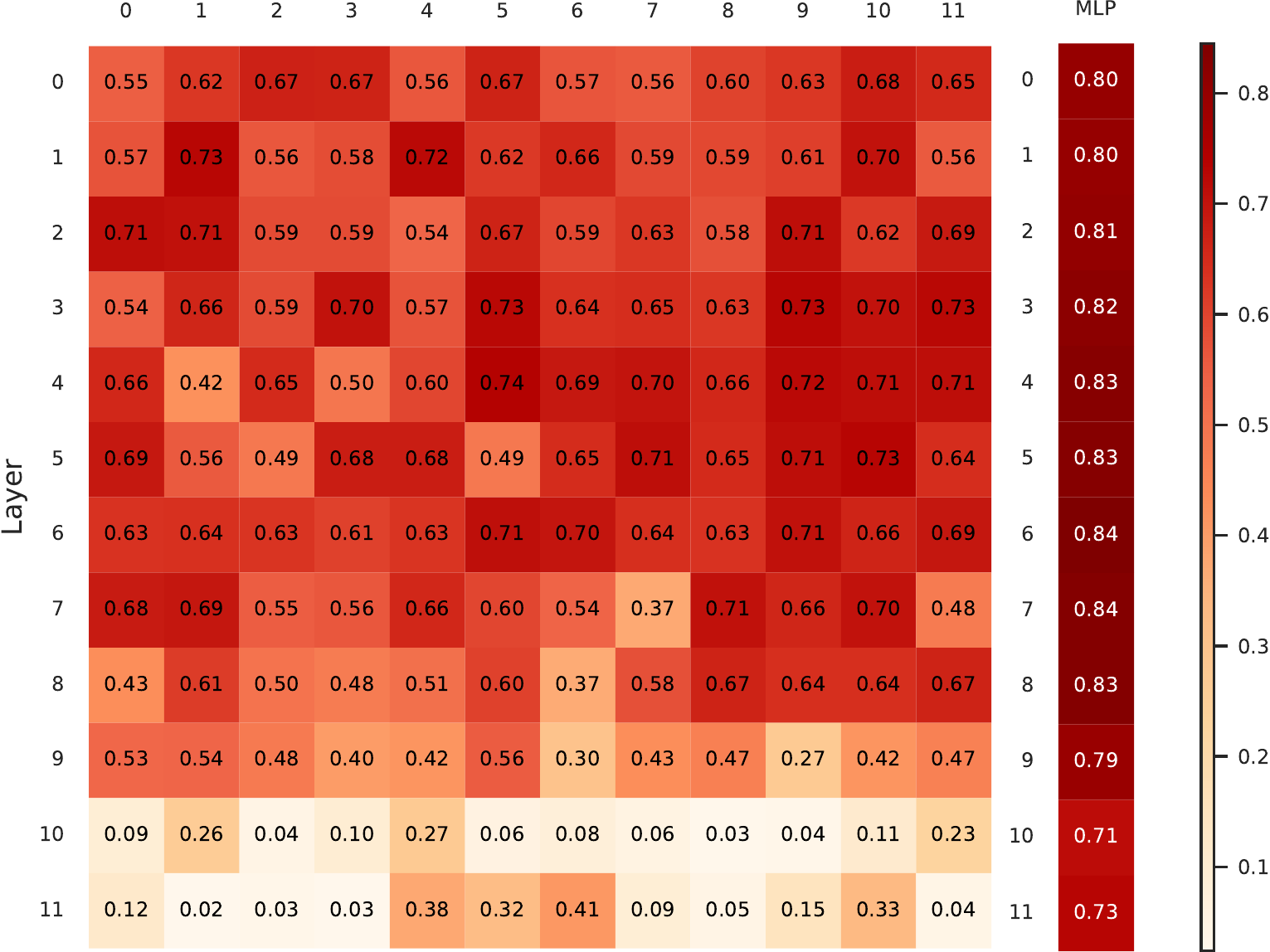}
}
\quad
\hspace{-0.2in}
\subfigure[SST-2]{
\label{fig:heatmap_mnli}
\includegraphics[width=5.1cm]{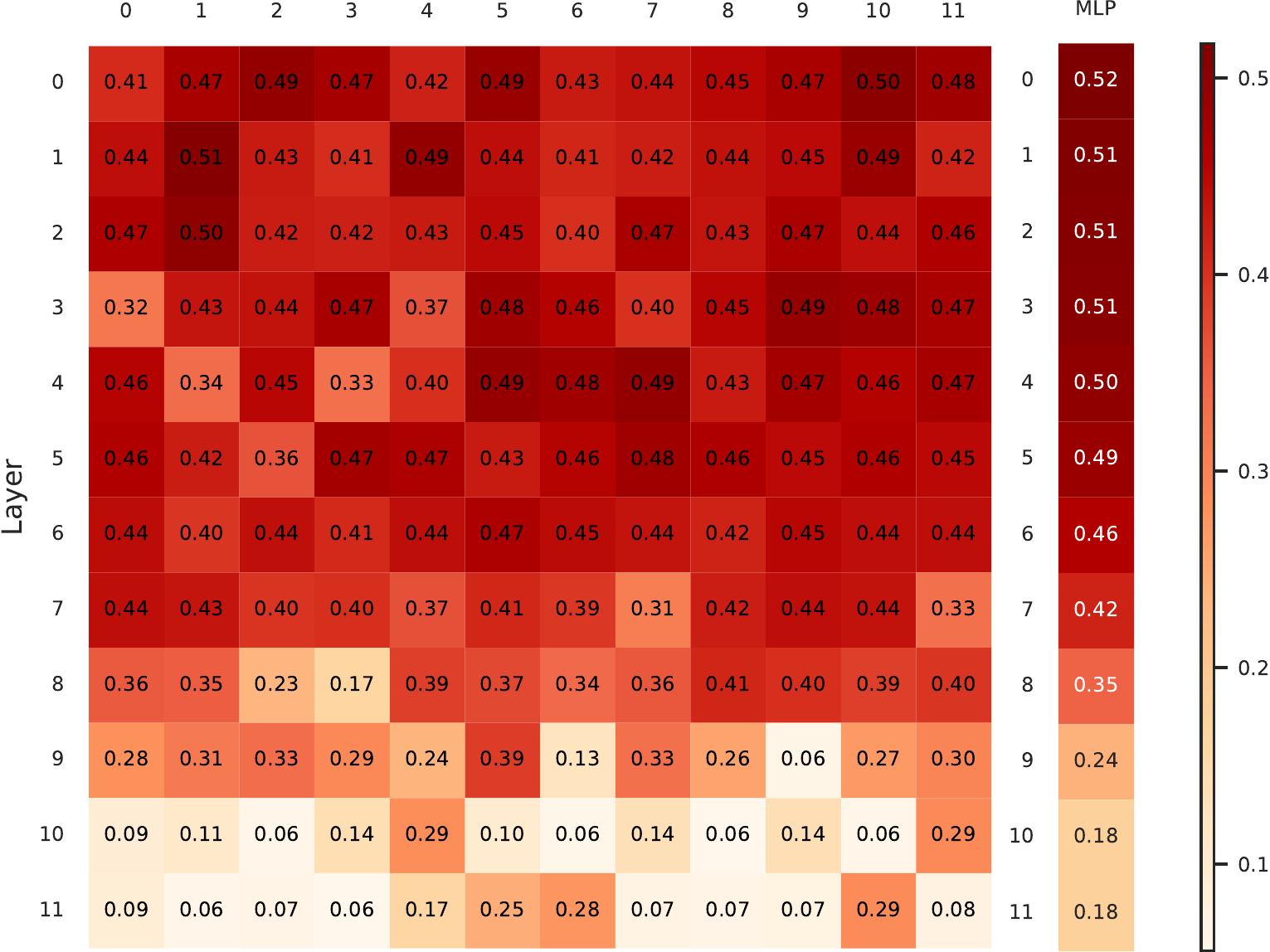}
}
\quad
\hspace{-0.2in}
\subfigure[AGNEWS]{
\label{fig:heatmap_mnli}
\includegraphics[width=5.1cm]{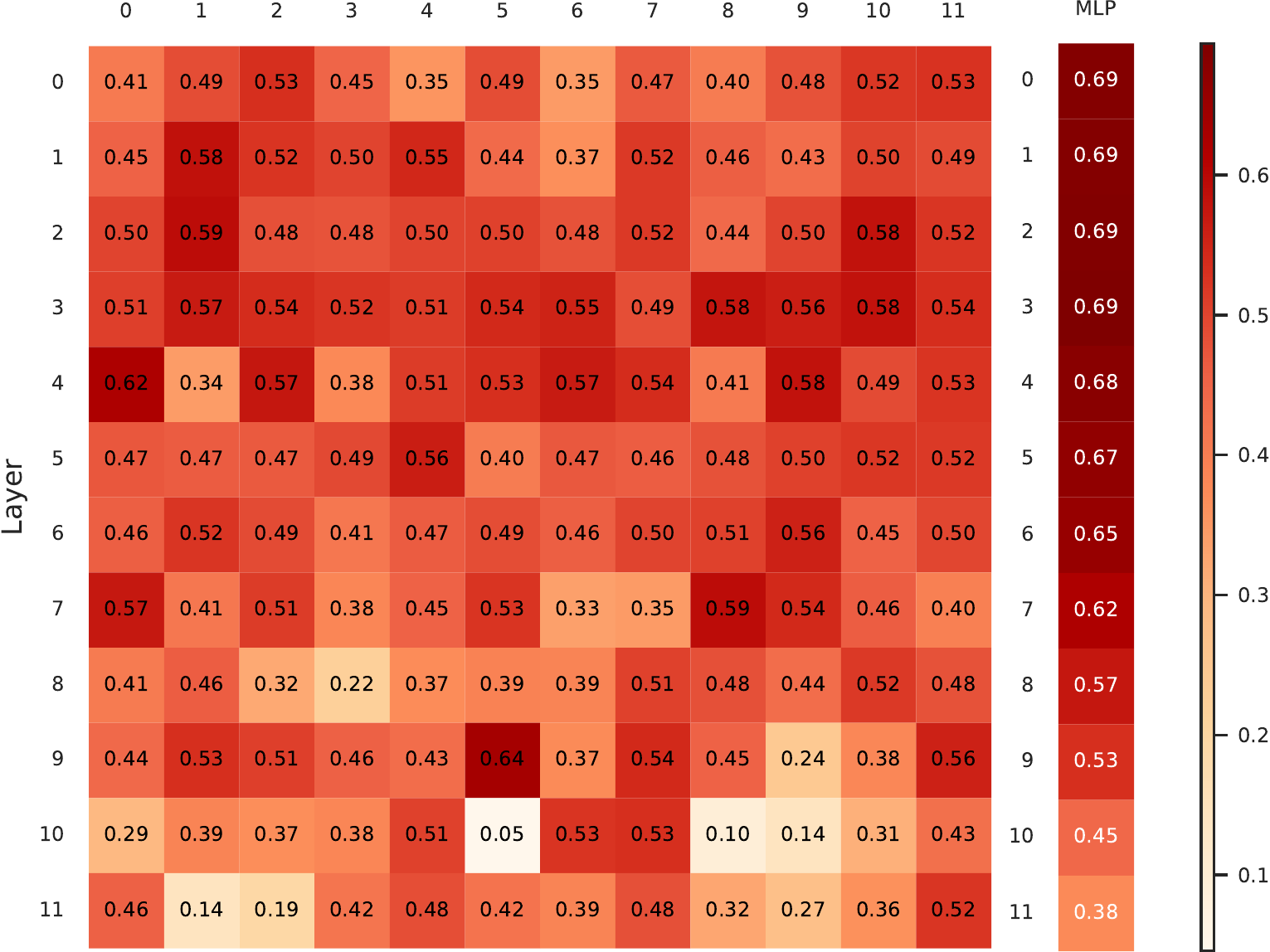}
}
\quad
\hspace{-0.2in}
\subfigure[MNLI]{
\label{fig:heatmap_mnli}
\includegraphics[width=5.1cm]{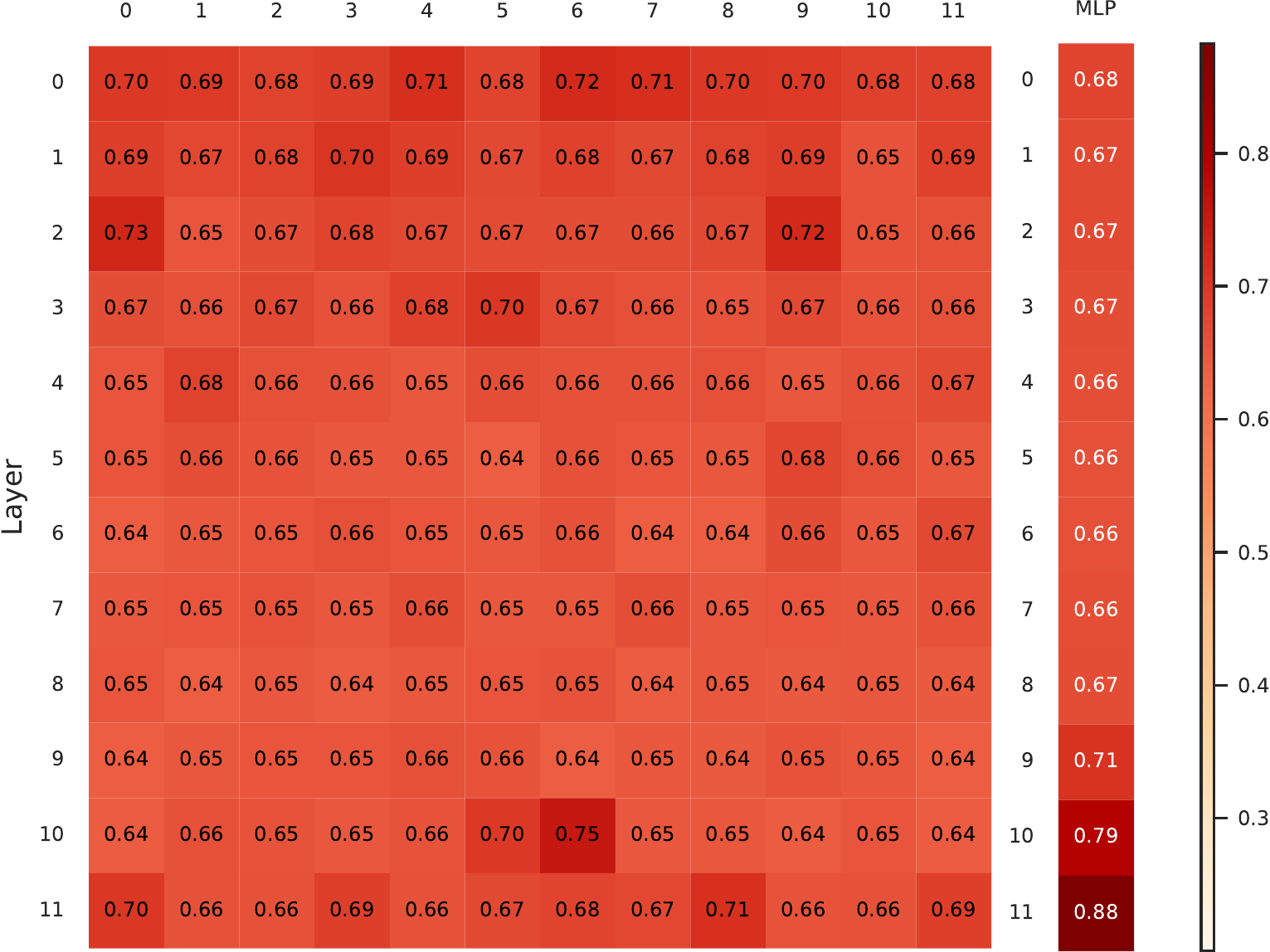}
}
\quad
\hspace{-0.2in}
\subfigure[QNLI]{
\label{fig:heatmap_mnli}
\includegraphics[width=5.1cm]{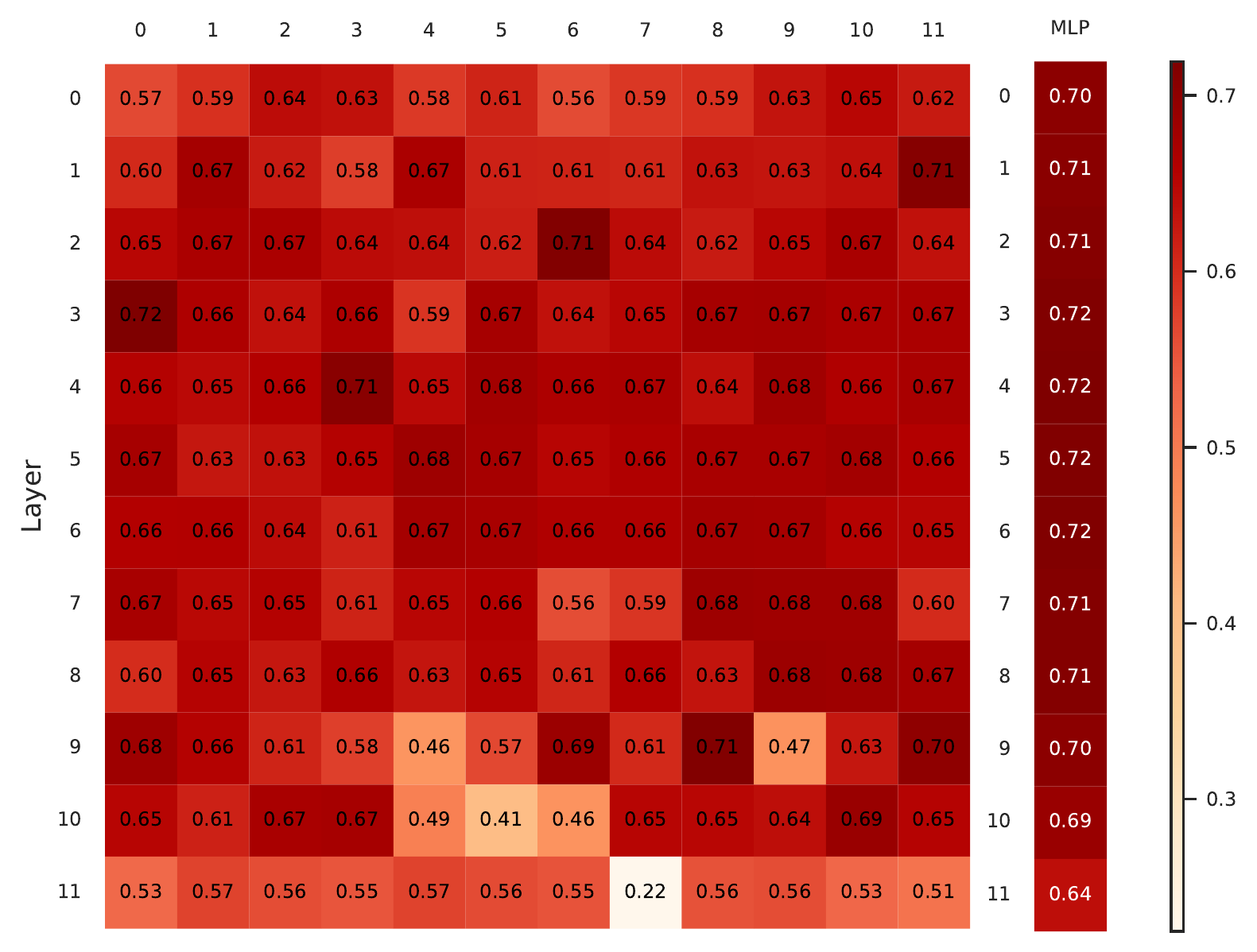}
}
\quad
\hspace{-0.2in}
\subfigure[QQP]{
\label{fig:heatmap_mnli}
\includegraphics[width=5.1cm]{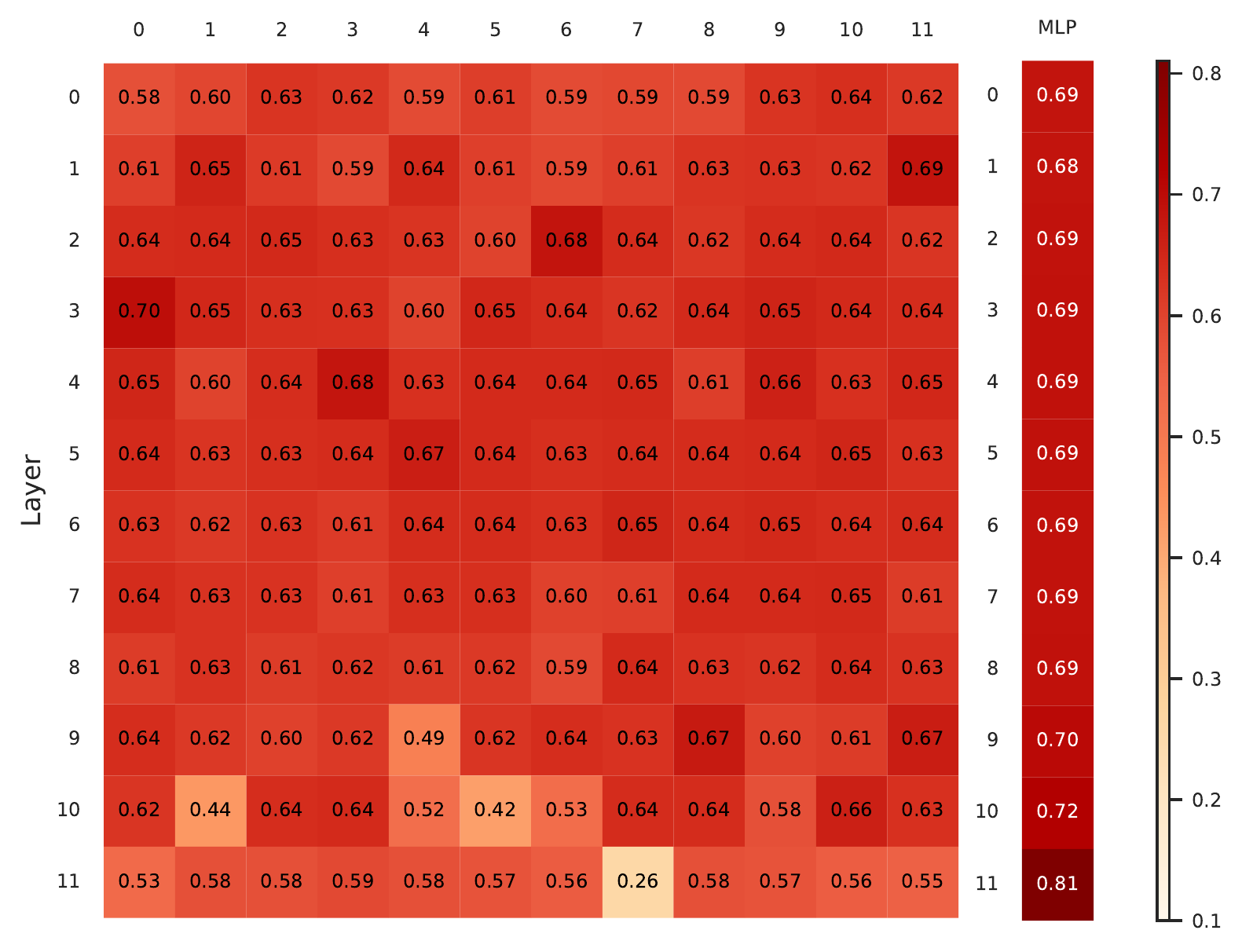}
}
\caption{Heatmaps of sparsity patterns found on different tasks, each cell gives the percentage of surviving weights in self-attention heads and MLPs. The sparsity patterns on IMDB and SST-2 are similar, which may be due to the fact that they are both text classification datasets based on movie reviews.}
\label{fig:heatmap_sparse}
\end{figure*}

The proposed method can prune out a subnetwork with arbitrary sparsity based on the confidence of masks.
In Fig.\ref{fig:sparse}, we compare the robust tickets and traditional 
lottery tickets across all sparsities. 
When the sparsity increases to a certain level, the robustness decreases faster than the accuracy, which indicates that the robustness is more likely to be affected by the model structure than the accuracy. Therefore, it is more difficult to find a robust ticket from BERT.
The accuracy of the subnetwork is slowly decreasing with increasing sparsity, but the robustness shows a different trend.
The change in robustness can be roughly divided into three phases: The robustness improves as the sparsity grows until a certain threshold; beyond this threshold, the robustness deteriorates but is still
better than that of the lottery tickets. In
the end, when being highly compressed, the robust network collapses into a lottery network.
A similar phenomenon is also be observed \cite{Liang2021SuperTI}.
The robustness performance curve is not as smooth as the accuracy, this may be due to the gap between the adversarial loss objective and the real textual attacks.

\subsection{Sparsity Pattern}

Fig.\ref{fig:heatmap_sparse} shows the sparsity patterns of robust tickets on all six datasets.
We can clearly find that the pruning rate increases from bottom to top on the text classification tasks (IMDB, SST2, AGNEWS), while it is more uniform in the natural language inference tasks (MNLI and QNLI) and Quora question pairs (QQP).
Recent works show that BERT encodes a rich hierarchy of linguistic information. 
Taking the advantage of the probing task, \citet{jawahar_what_2019} indicate that the surface information features are encoded at the bottom, syntactic information features are in the middle network, and semantic information features in the top.
Therefore, we speculate that the sparsity pattern of robust tickets is task-dependent.

\begin{figure*} [t]
\renewcommand\arraystretch{1.2}
\setlength\tabcolsep{5pt}
	\centering
	
	\subfigure[IMDB]{
		\centering
	\begin{minipage}[t]{0.32\textwidth}
	\includegraphics[width=5.2cm]{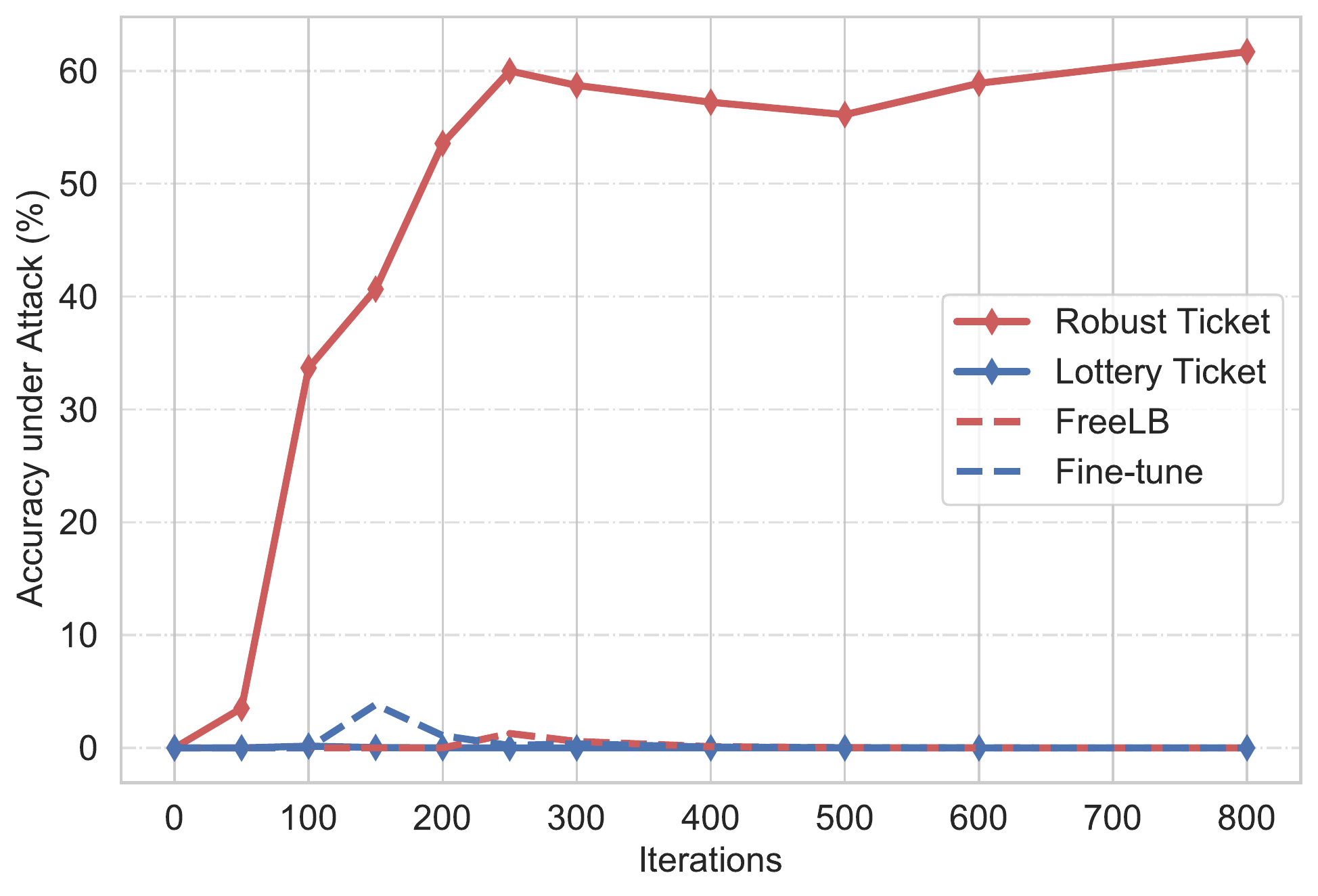}
	\includegraphics[width=5.2cm]{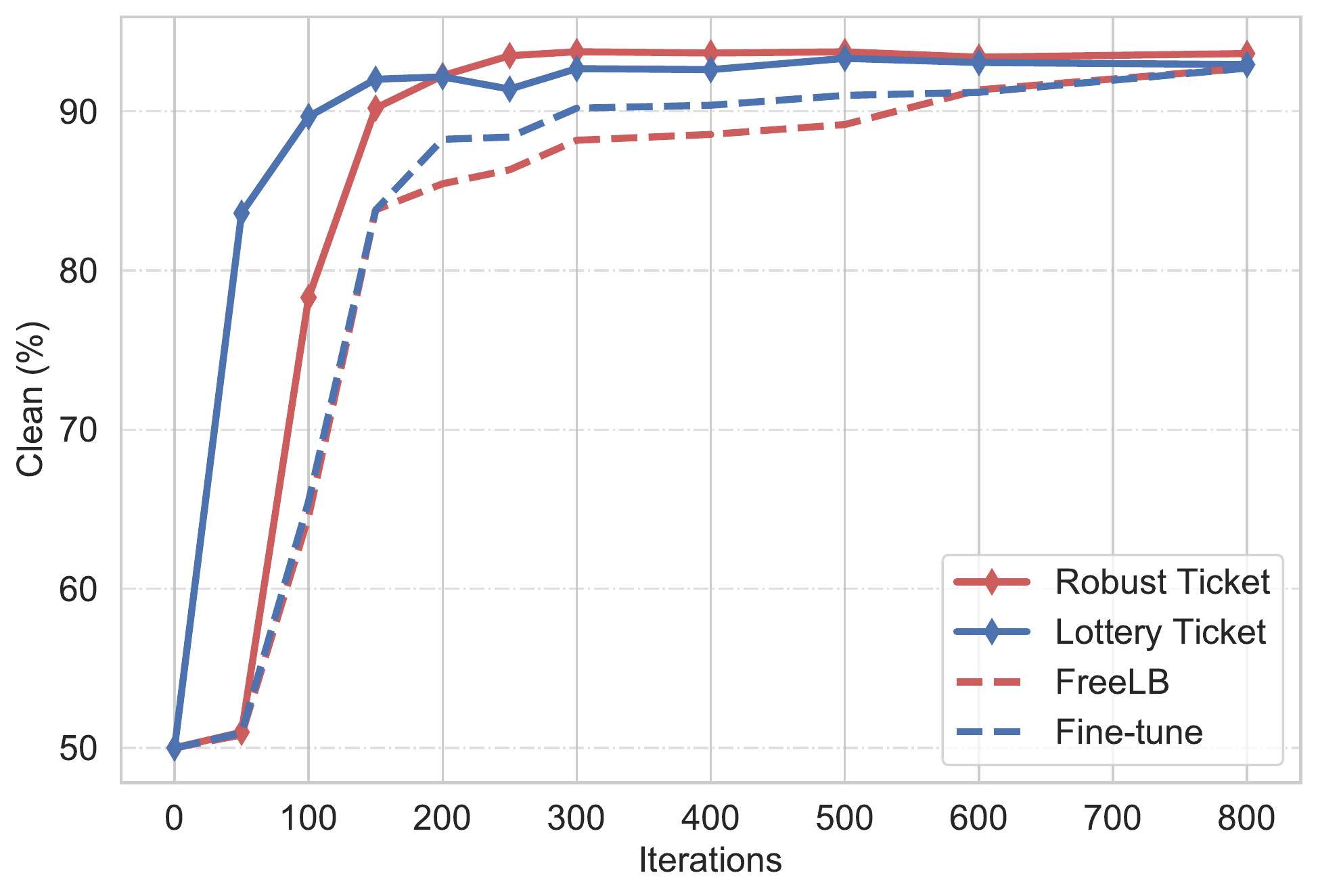}
	\end{minipage}
	}
	\subfigure[AGNEWS]{
		\centering
	\begin{minipage}[t]{0.3\textwidth}
		\includegraphics[width=5.2cm]{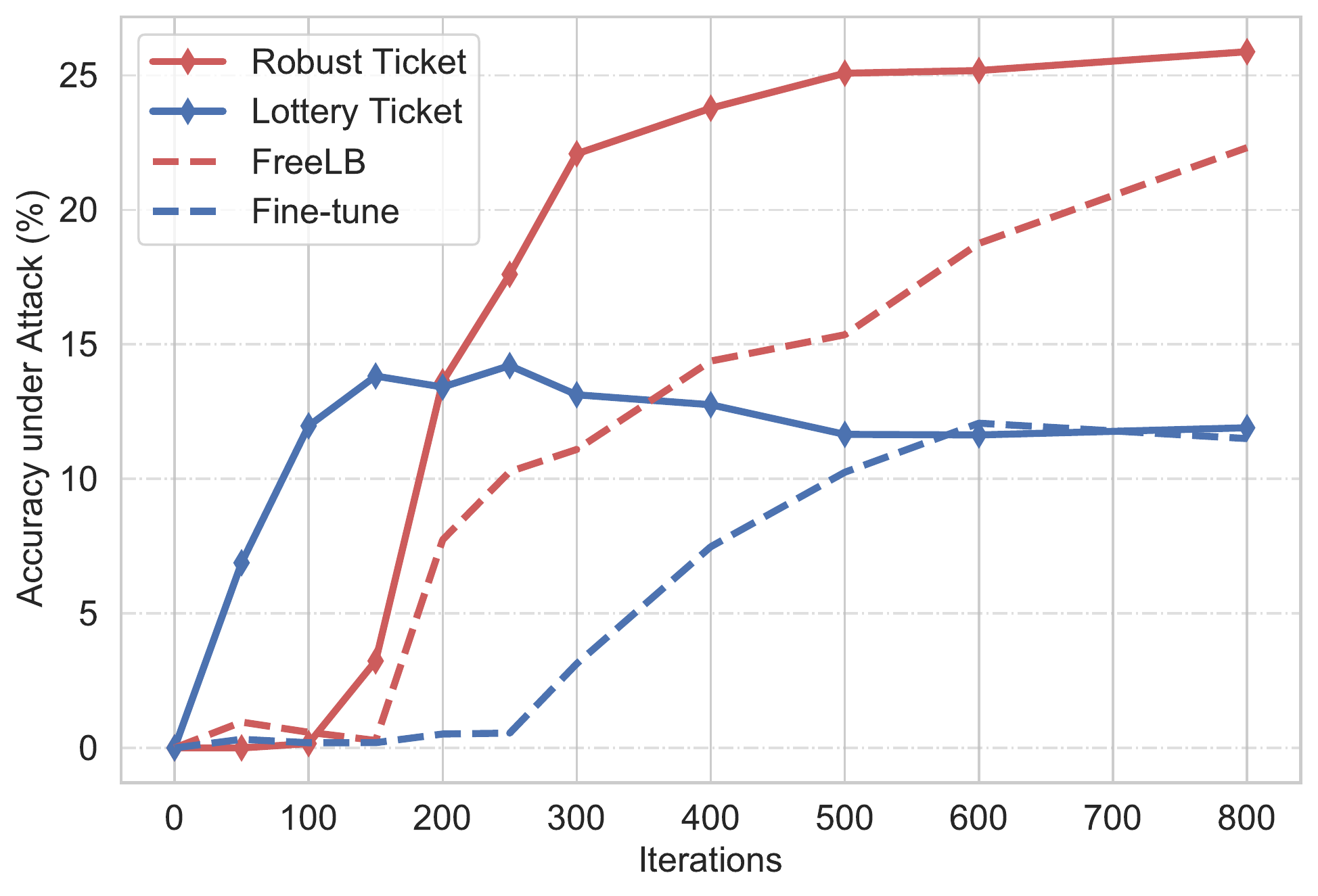}
		\includegraphics[width=5.2cm]{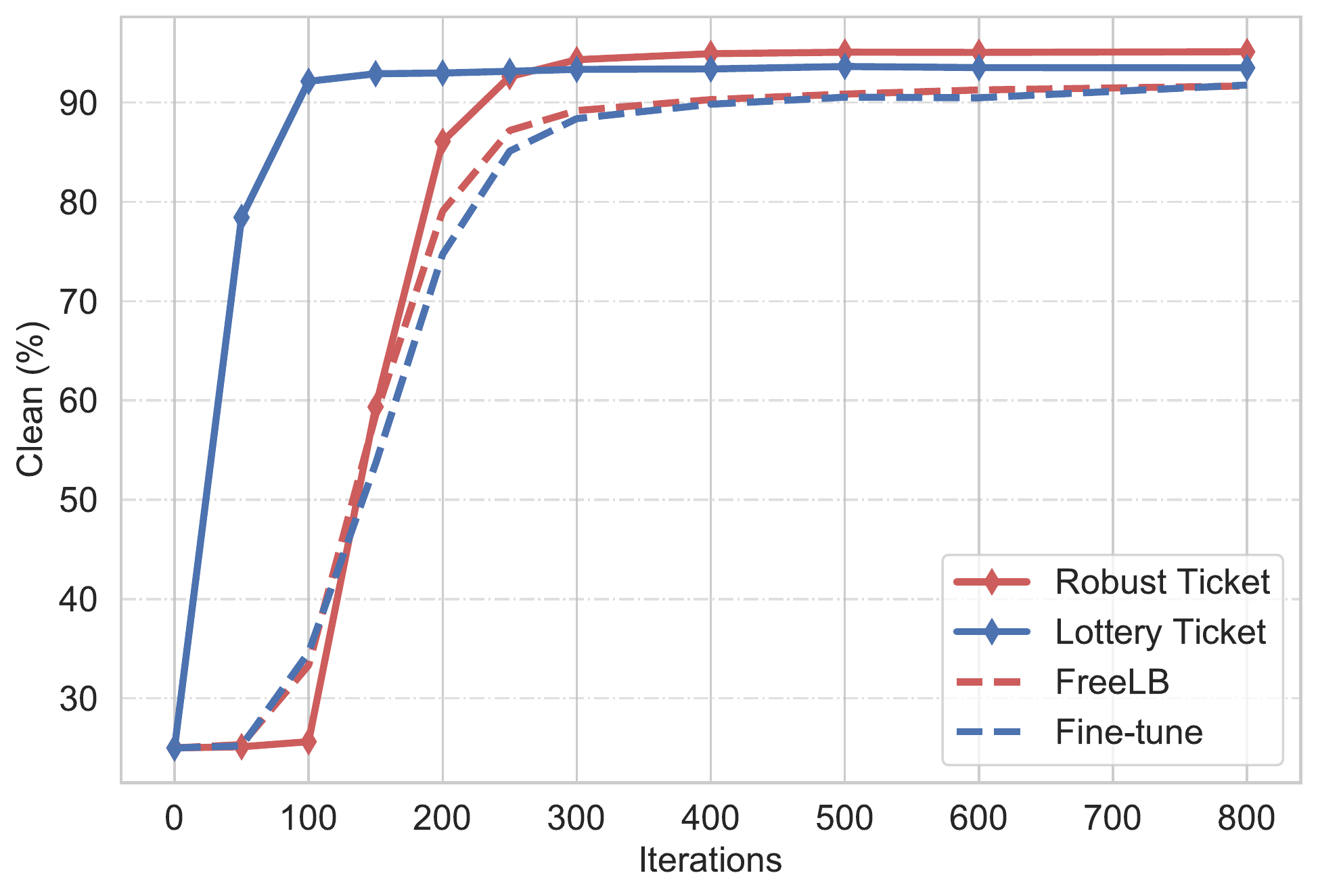}
	\end{minipage}
	}
	\subfigure[SST-2]{
		\centering
	\begin{minipage}[t]{0.32\textwidth}
		\includegraphics[width=5.2cm]{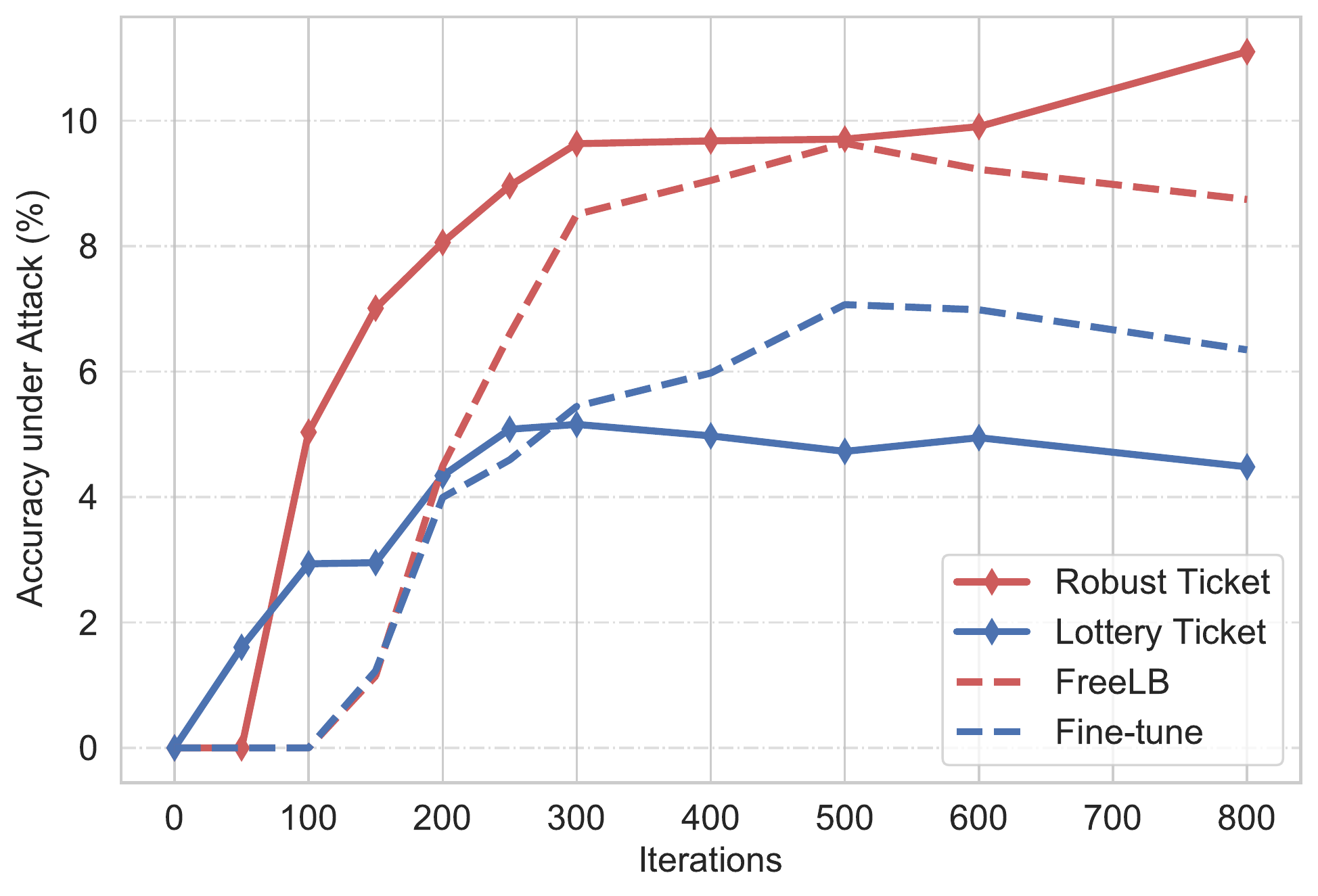}  	
		\includegraphics[width=5.2cm]{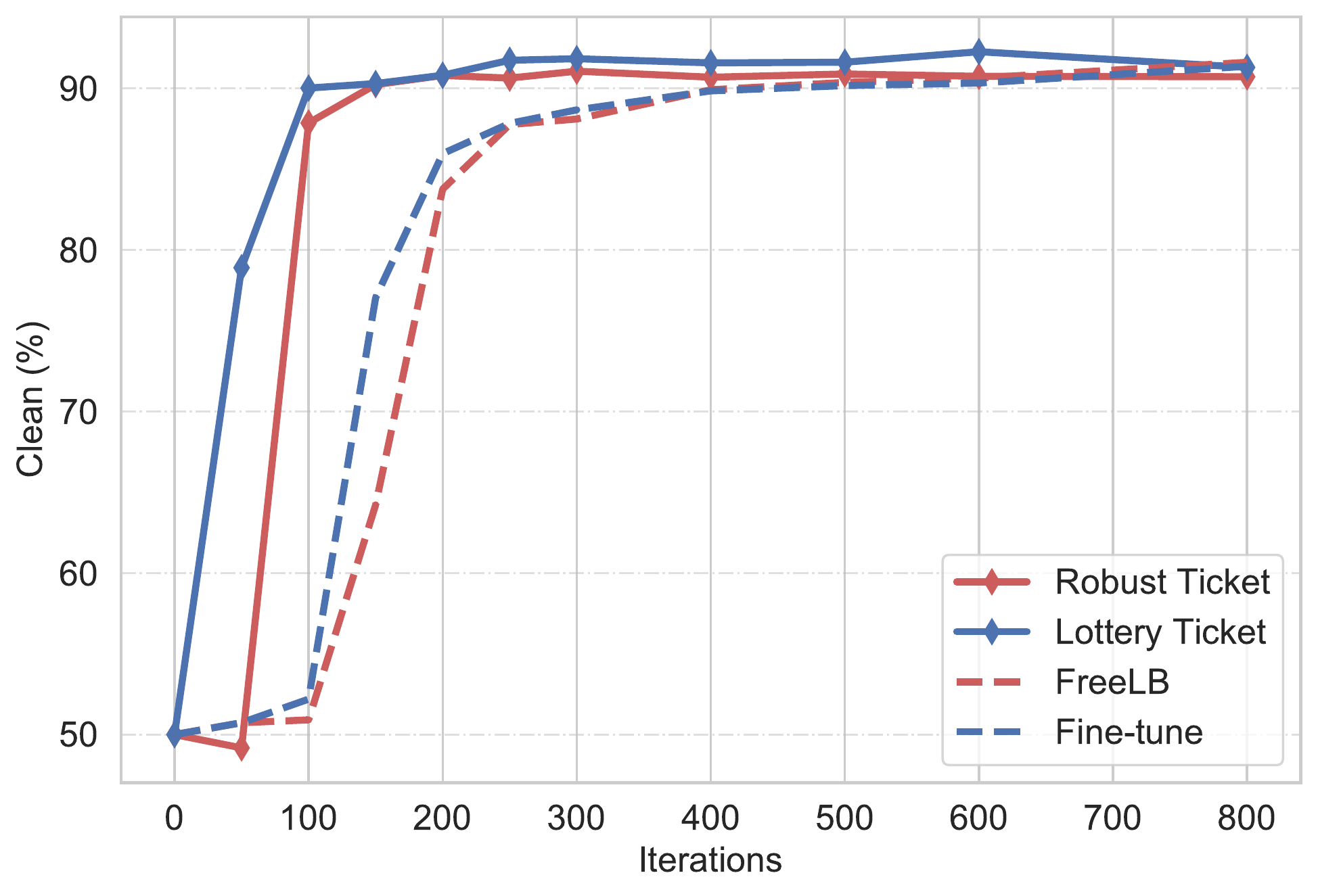}
	\end{minipage}
	}

\caption{Clean accuracy and accuracy under attack as training proceeds. 
Robust tickets accelerate both accuracy and robustness. 
\textbf{Aua$\%$} is obtained after using TextFooler attack.}
\label{fig:speed2}
\end{figure*}

\subsection{Speedup Training Process}
An important property of winning tickets is to accelerate the convergence of the training process \cite{Chen2021EarlyBERTEB, You2020DrawingET}. 
The training curve in Fig.\ref{fig:speed2} shows that the convergence speed of robust tickets is much faster compared with the default fine-tuning and FreeLB.
Moreover, the convergence rate of both accuracy and robustness is accelerating.
The traditional lottery tickets converge faster than our method, which may be due to the fact that robust tickets require maintaining a trade-off between robustness and accuracy.

\subsection{The Importance of Robust Tickets Initialization and Structure}

To better understand which factor, initialization or structure, has a greater impact on the robust ticket, we conduct corresponding analysis studies. 
We avoid the effect of initializations by re-initializing the weights of robust tickets. 
To avoid the effect of structures and preserve the effect of initializations, we use the full BERT and re-initialize the weights that are not contained in the robust tickets. 
\textbf{Aua$\%$} is obtained after using TextFooler attack.
The results are shown in Table \ref{tab:ablation}.

\begin{table}[t]
\renewcommand\arraystretch{1.4}
\setlength\tabcolsep{5pt}
\centering
\small
\begin{tabular}{l|lcc}
\hline
\hline
\textbf{Dataset}   & \multicolumn{1}{c}{\textbf{Method}} & \textbf{Clean$\%$} & \textbf{Aua$\%$} \\ \hline
 \multirow{4}{*}{$\textbf{IMDB}$}  & $\textbf{RobustT}_{20\%}$             & $\mathbf{93.7}$  & $\mathbf{55.6}$   \\
 & \; \textbf{w/o} Initialization   & $87.9$  & $0.2$   \\ 
 & \; \textbf{w/o} Structure                & $93.7$     & $13.4$   \\ 
 & \; \textbf{w/o} Structure+Longer               & $93.6$     & $18.6$   \\ \hline
\multirow{4}{*}{$\textbf{AGNEWS}$} &$ \textbf{RobustT}_{40\%}$            & $\mathbf{94.9}$     & $\mathbf{28.5}$   \\ 
& \; \textbf{w/o} Initialization                      & $92.4$     & $0.4$   \\ 
& \; \textbf{w/o} Structure                  & $94.9$     & $21.8$   \\ 
& \; \textbf{w/o} Structure+Longer                 & $94.8$     & $24.6$   \\ \hline
\multirow{4}{*}{$\textbf{SST-2}$}   & $\textbf{RobustT}_{30\%}$               &$90.9$     & $26.7$   \\ 
& \; \textbf{w/o} Initialization   & $83.1$ & $2.1$   \\ 
 & \; \textbf{w/o} Structure  & $\mathbf{92.0}$ & $15.7$  \\ 
& \; \textbf{w/o} Structure+Longer   & $91.9$ & $\mathbf{27.5}$ \\ \hline\hline

\end{tabular}
\caption{Importance of robust ticket initialization and structure. Our results show that the initialization of robust tickets seems to be more important than the structure, although both of them play a role.}
\label{tab:ablation}
\end{table}

\subsubsection{Importance of initialization}

LTH suggests that the winning tickets can not be learned effectively without its original initialization. 
For our robust BERT tickets, their initializations are pre-trained weights.
Table \ref{tab:ablation} shows the failure of robust tickets when the random re-initialization is performed.

\subsubsection{Importance of structure}

\citet{Frankle2019TheLT} hypothesize that the structure of winning tickets encodes an inductive bias customized for the learning task at hand. 
Although removing this inductive bias reduces performance compared to the robust tickets, it still outperforms the original BERT, and its performance improves further with longer training time (3 epochs $\rightarrow$ 10 epochs).
It can be seen that the initializations of some pre-training weights may lead to a decrease in the robustness of the model.

\section{Conclusion}

In this paper, we articulate and demonstrate the Robust Lottery Ticket Hypothesis for PLMs: the full PLM contains subnetworks (robust tickets) that can achieve a better robustness performance.
We propose an effective method to solve the ticket selection problem by encouraging weights that are not responsible for robustness to become exactly zero.
Experiments on various tasks corroborate the effectiveness of our method. 
We also find that pre-trained weights may be a key factor affecting the robustness on downstream tasks.

\section*{Acknowledgements}
The authors wish to thank the anonymous reviewers for their helpful comments. This work was partially funded by  National Natural Science Foundation of China (No. 62076069, 61976056). This research was supported by Meituan, Beijing Academy of Artificial Intelligence(BAAI), and CAAI-Huawei MindSpore Open Fund.

\bibliography{anthology,custom}
\bibliographystyle{acl_natbib}

\clearpage
\appendix

\section{The Effect of Regularization Strength during Mask Learning}\label{appendix0}

\begin{figure*}[t]
\subfigure[IMDB]{
\label{fig:sst2_sparse_aua}
\begin{minipage}[t] {0.35\linewidth}
\includegraphics[width=1.5\linewidth]{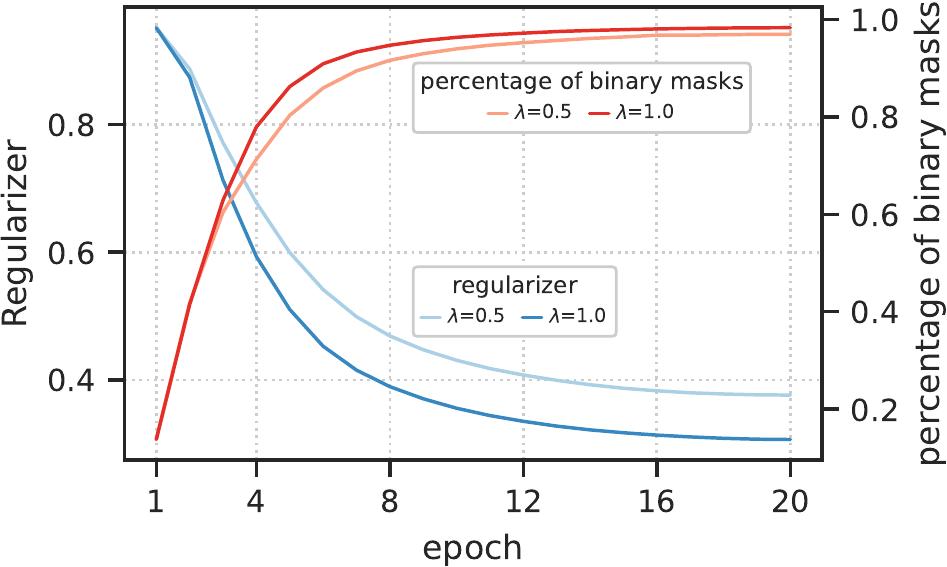}
\end{minipage}
}
\hspace{.6in}
\subfigure[AGNEWS]{
\label{fig:imdb_sparse_aua}
\begin{minipage}[t] {0.35\linewidth}
\includegraphics[width=1.5\linewidth]{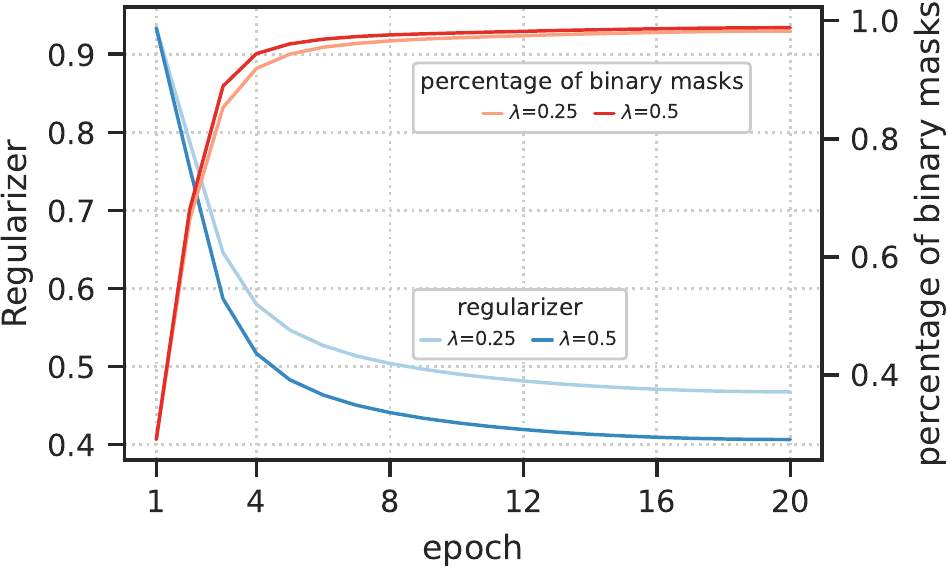}
\end{minipage}
}

\subfigure[QNLI]{
\label{fig:sst2_sparse_aua}
\begin{minipage}[t] {0.35\linewidth}
\includegraphics[width=1.5\linewidth]{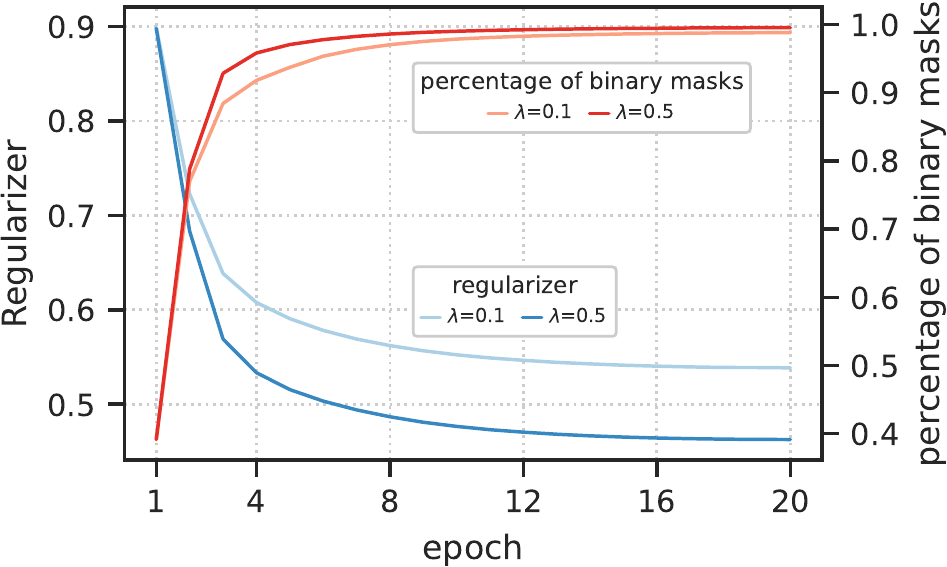}
\end{minipage}
}
\hspace{.6in}
\subfigure[QQP]{
\label{fig:imdb_sparse_aua}
\begin{minipage}[t] {0.35\linewidth}
\includegraphics[width=1.5\linewidth]{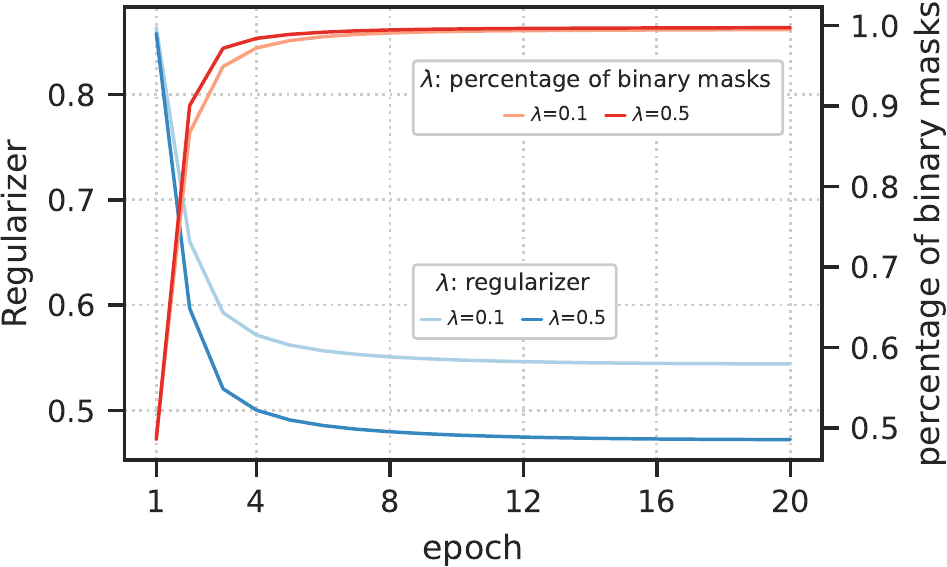}
\end{minipage}
}
\caption{Effect of regularization strength during mask learning.}
\label{fig:dot10}
\end{figure*}

In section \ref{effect_lambda}, we show the mask learning curves for various regularization strengths $\lambda$ in SST-2 dataset.
The results on more datasets are shown in the Fig.\ref{fig:dot10}, where we can observe that the mask learning process is insensitive to the regularization strength, and the convergence of masks is eventually achieved.

\section{Implementation Details}\label{appendix1}

\subsection{Details for Fine-tuning Models}

We report the hyperparameters used for fine-tuning the BERT-base and retraining the winning tickets in table \ref{tab:finetune-paramaters}.

\begin{table}[h]\small
\renewcommand\arraystretch{1.2}
\centering
\begin{tabular}{c|c}
\hline
\hline
\multicolumn{1}{c|}{\textbf{Hypeparameters}} &
\multicolumn{1}{c}{\textbf{Values}}\\
\hline
Optimizer & Adamw\cite{loshchilov2018decoupled} \\
Learning rate & $2\times10^{-5}$  \\ 
Dropout & $0.1$ \\
Weight decay & $1\times10^{-2}$ \\
Batch size & $16$ or $32$ \\
Gradient clip & $(-1,1)$ \\
Epochs & $3$ \\
Bias-correction & True \\
\hline
\hline
\end{tabular}
\caption{Hyperparameters used for fine-tuning the BERT-base and retraining the winning tickets.}
\label{tab:finetune-paramaters}
\end{table}

\subsection{Details for Adversarial Attack}\label{appendix2}
We use textattack \cite{morris2020textattack} to implement the adversarial attack methods.
For all attack methods, we use the default parameters of third-party libraries. 
Adversarial robustness evaluation metrics (e.g., \textbf{Aua$\%$} and \textbf{\#Query}) are evaluated on the all $872$ test samples for SST-2, $500$ randomly selected test samples for IMDB, and $1000$ randomly selected test samples for other datasets.

\subsection{Hyperparameters}\label{appendix3}
Adversarial loss objective introduces four widely used hyperparameters:
the perturbation step size $\eta$, the initial magnitude of perturbations $\epsilon_0$, the number of adversarial steps $s$, and we do not constrain the bound of perturbations.
In addition, we also report two important hyperparameters during mask learning.
They are mask learning rate $\gamma$ and regularization penalty coefficient $\lambda$.
The weight decay $wd$ in the optimizer are also changed compared with default
settings to make the mask sparsity rate converge better. 
We list the hyperparameters used for each tasks in Table \ref{tab:mask-finding-args}.

\begin{table}[t]
\renewcommand\arraystretch{1.2}
\centering
\small
\begin{tabular}{c|cccccc}
\hline
\hline
\multicolumn{1}{c|}{\textbf{Datasets}} &
\multicolumn{1}{c}{\textbf{$\eta$}} &
\multicolumn{1}{c}{\textbf{$\gamma$}} &
\multicolumn{1}{c}{\textbf{$\lambda$}} &
\multicolumn{1}{c}{\textbf{$\epsilon_0$}} &
\multicolumn{1}{c}{\textbf{$s$}} &
\multicolumn{1}{c}{\textbf{$wd$}}\\
\hline
\textbf{SST2} & $0.03$ &$0.1$& $0.5$& $0.05$& $5$ & $1e-6$\\
\textbf{AGNEWS} & $0.03$ &$0.05$& $0.5$& $0.05$& $5$ & $1e-6$\\
\textbf{IMDB} & $0.03$ &$0.1$& $0.5$& $0.05$& $5$ & $1e-6$\\
\textbf{QQP} & $0.04$ &$0.05$& $0.1$& $0.05$& $3$ & $1e-6$\\
\textbf{QNLI} & $0.04$ &$0.05$& $0.1$& $0.05$& $3$ & $1e-6$\\
\textbf{MNLI} & $0.2$ &$0.1$& $0.1$& $0.05$& $2$ & $1e-6$\\

\hline
\hline
\end{tabular}
\caption{Hyperparameters used during mask learning.}
\label{tab:mask-finding-args}
\end{table}

\end{document}

%% file: definition.tex
\def\<{\left\langle }
\def\>{\right\rangle }

\def\digamma{{\mathrm{digamma}}}